\documentclass{article} % For LaTeX2e
\usepackage{salesforce,times}

\usepackage{amsmath,amsfonts,bm}

\def\eqref#1{equation~\ref{#1}}
\def\1{\bm{1}}

\DeclareMathAlphabet{\mathsfit}{\encodingdefault}{\sfdefault}{m}{sl}
\SetMathAlphabet{\mathsfit}{bold}{\encodingdefault}{\sfdefault}{bx}{n}

\usepackage[ruled,vlined]{algorithm2e}
\usepackage{wrapfig}
\usepackage{xcolor}
\usepackage{hyperref}
\usepackage{url}

\definecolor{hlred}{RGB}{200,0,0}

\usepackage{graphicx}
\usepackage{booktabs} 
\usepackage{array} 
\usepackage{enumitem}
\usepackage[most,skins,theorems]{tcolorbox}
\definecolor{aiboxblue}{RGB}{225,238,252} 
\tcbset{
  aibox/.style={
    width=\linewidth,
    top=8pt,
    bottom=4pt,
    colback=aiboxblue,
    colframe=black,
    colbacktitle=black,
    enhanced,
    center,
    attach boxed title to top left={yshift=-0.1in,xshift=0.15in},
    boxed title style={boxrule=0pt,colframe=white,},
  }
}
\newtcolorbox{AIbox}[2][]{aibox,title=#2,#1}

\newcommand{\methodlong}{Reward-Integrity Verified Environments for RL}
\newcommand{\method}{\textsc{River}}

\title{Learning Generalizable Behaviors for \\Terminal Agents}

\author{
Yihang Yao$^{1,2,\dagger}$, Bo Pang$^{1,\dagger}$, Xuan Phi Nguyen$^{1}$, Ding Zhao$^{2}$, \textbf{Shafiq Joty}$^{1}$, \textbf{Semih Yavuz}$^{1}$
}
\sfsetaffiliation{$^1$Salesforce AI Research, $^2$Carnegie Mellon University; $\dagger$ Core contributors.}
\sfsetauthornote{~~Contact: \texttt{yihangya@andrew.cmu.edu}}

\begin{document}

\maketitle

\begin{abstract}

% \shafiq{This is my version. Old version is commented out below} 

Terminal agents have emerged as a compelling application of large language models (LLMs), with the potential to integrate deeply into users' daily workflows. Reinforcement learning (RL) has become a key technique for improving their capabilities, making the scalability of training environments a central challenge. Since public real-user interaction data are scarce, synthetic environments provide a practical alternative. However, they often exhibit substantial domain gaps and limited fidelity, resulting in poor generalization. Existing efforts primarily focus on scaling the quantity and diversity of synthetic environments, while the reward signal quality is under-explored and the mechanisms that govern generalization remain poorly understood.

% {In this work, we study how RL improves terminal agents and propose the idea of \textbf{agentic compositional generalization}: rather than learning domain-specific skills from scratch, RL primarily acquires high-level decision-making behaviors that compose and route the low-level skills already learned during pre-training and supervised fine-tuning (SFT), enabling strong out-of-distribution (OOD) generalization. This perspective suggests that verifier quality, which shapes these behaviors, is more important than simply increasing the number or diversity of training environments.} 

In this work, we study how RL improves terminal agents and propose the \textbf{agentic compositional generalization} hypothesis: rather than teaching the agent new domain-specific skills from scratch, RL primarily shapes high-level decision-making behaviors that compose and route the low-level skills already acquired during pre-training and supervised fine-tuning (SFT). This account is consistent with the evaluation performance we observe.
This perspective suggests that verifier quality, which determines which behaviors are reinforced by providing optimization signals in RL, 
is more important than simply increasing the number or diversity of training environments. 
Motivated by this insight, we propose \method, a simple training recipe that improves the reward signal by filtering low-quality environments and augmenting the outcome reward with process-level behavior regularization.
Using this recipe, our RL-trained agent achieves the best performance among evaluated open-source RL-trained 8B models across four terminal-agent benchmarks. Moreover, our approach generalizes across model families, scales, agent harnesses, and RL objectives. Using \textbf{fewer than 30\%} of the TMax training environments, \method\ improves RL gains by \textbf{106\% and 30\%} on average for models ranging from 2B to 27B on Terminal-Bench-Lite and Terminal-Bench-v2.1, respectively. Our Website: \href{https://self-evolving-agents.salesforceresearch.ai/papers/river/index.html}{\textcolor{blue}{https://self-evolving-agents.salesforceresearch.ai/papers/river/index.html}}.

\end{abstract}

\section{Introduction}
% \shafiq{This is my version. Old version is commented out below} 

As large language models (LLMs) continue to advance toward real-world deployment, \emph{terminal agents} have emerged as one of their most important applications, enabling software engineering, system administration, and other interactive coding tasks~\citep{anthropic2025claudecode, openai2025codex, pi2026data, xai2025grokcode, ivison2026tmax, wu2026swe}. While the capabilities of frontier terminal agents have improved rapidly, the principles underlying the design of their training environments remain largely undisclosed~\citep{pi2026data, ivison2026tmax, zhu2026termigen}. In particular, it remains unclear how post-training for terminal agents generalizes across domains, and what factors matter beyond simply increasing coverage over task categories and targeted skills. 

Reinforcement learning (RL) has become the dominant paradigm for post-training reasoning models and agents~\citep{openai2024o1card, guo2025deepseek}. Unlike single-turn reasoning tasks such as mathematical reasoning~\citep{shao2024deepseekmath}, 
terminal agents require interacting with executable environments through long-horizon, multi-turn decision making~\citep{gandhi2026endless}. Consequently, the training environment plays a central role in agentic RL. Since large-scale real-user interaction data are generally unavailable, recent work has increasingly relied on \emph{synthetic environments} as a scalable alternative~\citep{pi2026data, li2026recursive}.

Existing work on synthetic terminal environments has primarily focused on \emph{scaling coverage}: expanding the diversity of tasks, domains, and required skills~\citep{pi2026data, fan2026toward, zhao2026nexforge}. Early efforts generated demonstration trajectories for supervised fine-tuning (SFT), whereas more recent work constructs executable environments for RL. For example, Endless-Terminals~\citep{gandhi2026endless} demonstrated that RL over synthetic environments substantially improves terminal agents, and TMax~\citep{ivison2026tmax} further increased environment diversity and domain coverage. However, \emph{coverage alone does not explain generalization}. This raises two fundamental questions: \textit{(i)} What does RL actually learn during terminal-agent training? \textit{(ii)} What makes one training environment more valuable than another, aside from diversity, for improving generalization beyond the synthetic training-task distribution?

In this work, we study these questions through the lens of the \emph{compositional generalization} hypothesis \citep{yuan2026f, kong2026reasoning}, extending it from single-turn reasoning models to multi-turn long-horizon agentic models. We term the resulting account \textit{agentic compositional generalization} (see Fig.~\ref{fig:teaser-figure-triple} for an overview), and test it in realistic terminal environments rather than purely synthetic ones. 
Our findings indicate complementary roles of SFT and RL in generalization: SFT (and general model prior imposed by pre-training and mid-training) primarily equips the agent with low-level atomic skills, while RL primarily shapes \emph{high-level decision-making behaviors} that compose, route, and coordinate these skills across domains. Under this hypothesis, generalization requires both coverage on atomic skills and reusable high-level behaviors.

This perspective suggests a different axis for the design of the training recipe. 
% For behaviors, if RL is not efficient in learning atomic skills, then SFT should equip the models with diverse atomic skills that are absent from model priors.
% If RL primarily learns high-level behaviors, then \emph{verifiers} in synthetic environments, which provide the signal that shapes those behaviors, becomes important. Guided by this insight, we propose \method, a simple yet effective training recipe that filters out low-quality environments and improves verifier quality through behavior-level regularization. Despite using fewer than $30\%$ of the TMax training environments, \method\ consistently improves RL efficiency and generalization across model families, model scales, agent harnesses, and RL algorithms.
For skills, since RL is not efficient in learning atomic skills, then SFT should equip the models with diverse atomic skills that are absent from model priors.
For behaviors, since RL primarily learns high-level behaviors, then \emph{verifiers} in synthetic environments, which provide the signal that shapes those behaviors, become critical. Guided by this insight, we propose \method, a simple yet effective training recipe that filters out low-quality environments and improves verifier quality through behavior-level regularization. Despite using fewer than $30\%$ of the TMax training environments, \method\ consistently improves RL efficiency and generalization across model families, model scales, agent harnesses, and RL algorithms.

Our contributions are summarized as follows:

\begin{enumerate}[leftmargin=*,label=(\arabic*)]
    \item \textbf{Mechanistic interpretation.}
    We propose \emph{agentic compositional generalization}, which posits that RL improves generalization by shaping reusable high-level behaviors that compose pre-RL domain atomic skills, and we present empirical evidence consistent with its predictions.

    \item \textbf{A simple and effective training recipe.}
    We introduce \method, which improves the quality of reward signal and the training efficiency. Using fewer than $30\%$ of the TMax environments~\citep{ivison2026tmax}, \method\ achieves on average $106\%$ and $30\%$ larger RL gains on Terminal-Bench-Lite and Terminal-Bench-v2.1, respectively, across models ranging from $2$B to $27$B.

    \item \textbf{Comprehensive empirical analysis.}
    We provide extensive experiments that test the predictions of our hypothesis and offer new insights into the learning dynamics of RL for terminal agents.
\end{enumerate}

\section{Related Work}
%\textbf{Terminal Agents} such as Claude Code~\citep{anthropic2025claudecode} and Codex~\citep{openai2025codex} have been widely used in software engineering tasks and evaluated on long-horizon coding benchmarks~\citep{badertdinov2026swe, qiu2025locobench}, including Terminal-Bench~\citep{merrill2026terminal} and SWE-Bench~\citep{jimenez2024swe}. The open-source and academic communities are also actively exploring recipes for training terminal agents~\citep{wang2025let, zeng2025satori, guha2026openthoughts, raoof2026openthoughts, peng2025litecoder, cheng2026terminal, bercovich2026terminal}. RL, which has proven effective for training long-horizon agents~\citep{yang2026gui, wang2026openclaw}, has also been applied to terminal agents. Among these efforts, scaling high-quality data and environments has become one of the core challenges~\citep{pi2026data, ivison2026tmax, ruan2026evolvecoder}.

% \shafiq{This is my version. Old version is commented out below} 

\textbf{Terminal Agents} such as Claude Code~\citep{anthropic2025claudecode} and Codex~\citep{openai2025codex} have rapidly become central to software-engineering workflows, demonstrating strong performance on long-horizon coding benchmarks such as Terminal-Bench~\citep{merrill2026terminal}, SWE-Bench~\citep{jimenez2024swe}, and their variants~\citep{badertdinov2026swe, qiu2025locobench}. Motivated by these advances, the open-source and research communities have increasingly explored recipes for training terminal and coding agents~\citep{wang2025let, zeng2025satori, guha2026openthoughts, raoof2026openthoughts, peng2025litecoder, cheng2026terminal, bercovich2026terminal}. Among the various post-training approaches, reinforcement learning (RL) has emerged as one of the most effective techniques for improving long-horizon agentic behaviors~\citep{yang2026gui, wang2026openclaw}, and has recently been adopted for terminal-agent training. A key challenge, however, is scaling high-quality training data and executable environments, which has become one of the primary bottlenecks for further progress~\citep{pi2026data, ivison2026tmax, ruan2026evolvecoder}.

\textbf{Synthetic Data and Environment Scaling.} Scaling synthetic data has become a central paradigm for training LLM agents, driven by the scarcity and cost of collecting real-world interaction data~\citep{chen2026scaling, zhang2025agent, cen2026webscale}. Recent efforts have evolved from generating static demonstrations for supervised fine-tuning (SFT)~\citep{liu2024apigen, xu2025toucan, prabhakar2026apigen} to constructing interactive training environments~\citep{yang2026swe, zhou2026synthetic, lin2026cli, li2026recursive, he2026frontiersmith}, which provide the executable feedback required for large-scale reinforcement learning. For terminal agents in particular, most existing work has focused on expanding environment coverage across skills, domains, and task categories~\citep{ivison2026tmax, pi2026data}. While recent studies have shown that verifier quality, such as stronger test cases, can improve coding performance~\citep{he2025hardtests}, the broader role of verifier quality in shaping learned behaviors and enabling out-of-distribution generalization for RL-trained terminal agents remains largely unexplored.

%\textbf{Synthetic Data and Environment Scaling} has been widely studied in LLM-agent training, as real-world data is expensive and hard to collect at scale. Recent work has shifted from scaling static demonstration data for SFT~\citep{liu2024apigen, xu2025toucan, prabhakar2026apigen} to building interactive environments~\citep{yang2026swe, zhou2026synthetic, lin2026cli, li2026recursive, he2026frontiersmith}, which are better suited to RL at scale. For terminal agents specifically, most work focuses on coverage over skills and task categories~\citep{ivison2026tmax, pi2026data}. Although test-case quality has been shown to matter for coding agents~\citep{he2025hardtests}, how verifier quality affects the generalization of trained terminal agents remains underexplored.

\textbf{Learning Dynamics and Generalization Mechanisms.} A growing body of work seeks to understand why LLMs and agents generalize beyond their training distributions and how these insights can guide more effective training recipes~\citep{chu2025sft, huang2025self, ren2025learning, chen2026coverage, YXLA2024-gsm1, yu2025demystifying, rajaraman2026learning, yue2504does}. Recent works on \emph{compositional generalization}~\citep{kong2026reasoning, yuan2026f} explain the complementary roles of SFT and RL in single-turn reasoning models: SFT provides the low-level reasoning primitives, while RL learns to compose and route these primitives to enable generalization. Other studies investigate reasoning behaviors and decision-making patterns, showing how understanding these dynamics can improve RL training~\citep{gandhi2025cognitive, cen2026behavior, yeo2025demystifying, yao2026tailored}, with recent extensions to multi-turn agentic settings~\citep{yao2026pushing}. A complementary line of work examines how spurious or misaligned reward signals affect RL learning and generalization~\citep{shao2025spurious}.
% In contrast, we study the learning dynamics of \emph{terminal agents} operating in realistic interactive environments. We extend the compositional generalization hypothesis to the multi-turn agentic setting and show how the resulting insights can guide the design of more effective RL training environments and recipes.
In contrast, we study the learning dynamics of \emph{terminal agents} operating in realistic interactive environments. We extend the compositional generalization hypothesis to the multi-turn agentic setting, quantitatively dissociate single-turn skills from multi-turn behaviors at the feature level (see Sec.~\ref{section: analysis}), and show how the identified behaviors can be targeted directly through advantage shaping and used to guide the design of RL training environments and recipes.

%\textbf{Learning Dynamics and Generalization Mechanisms.} Many recent works explain how LLMs and agents generalize beyond their training domains and derives insights for designing training recipes~\citep{ren2025learning, YXLA2024-gsm1}. Compositional generalization~\citep{kong2026reasoning, yuan2026f} explains the roles of SFT and RL in single-turn reasoning models: SFT supplies the raw module materials in compositional traces, while RL decomposes those traces to identify the latent atomic modules and enable compositional generalization. Other work studies reasoning patterns and how to exploit them to improve RL performance~\citep{gandhi2025cognitive, cen2026behavior, yao2026tailored}, and recent work extends such patterns to multi-turn settings~\citep{yao2026pushing}. A further line of work examines the effects of spurious rewards~\citep{shao2025spurious}. In this work, we focus on understanding the generalization mechanism of agentic models in the real-world terminal setting and use the resulting insights to guide recipe design.

% Behavior analysis: Reasoning models~\citep{gandhi2025cognitive, cen2026behavior}, Agentic behavior~\citep{yao2026pushing, }

% PACE paper~\citep{song2026pace}

% \clearpage
% \newpage
\section{Mechanism Interpretation and Method}
\label{section:method}
% \shafiq{This is my version; old version is commented out}

Synthetic environments typically introduce a substantial training-evaluation distribution gap: evaluation tasks are often significantly more complex than those posed by the synthetic training environments, and no practical training corpus can exhaustively cover the diversity of real-world task categories.
In this section, we first present an \textbf{Agentic Compositional Generalization} hypothesis, our account of how RL-trained terminal agents achieve generalization beyond the synthetic environment distribution and close the training-evaluation gap. 
Guided by this perspective, we then introduce Reward-\textbf{I}ntegrity \textbf{V}erified \textbf{E}nvironments for \textbf{R}L (\textbf{\method}), a simple training pipeline that improves the quality of synthetic environments through verifier-aware filtering and behavior-level regularization, and we describe the resulting RL training recipe.

%Synthetic environments introduce a large training-evaluation gap in data distribution: Evaluation tasks are usually much more complex than training data, and training environments can hardly cover all the evaluation task categories. In this section, we first propose the hypothesis, \textbf{Agentic Compositional Generalization}, explaining how RL-trained terminal agents achieve OOD generalization, and then introduce the pipeline, \textbf{R}eward-\textbf{I}ntegrity \textbf{V}erified \textbf{E}nvironments for \textbf{R}L~(\textbf{\method}), we use to improve the quality of synthetic environments, together with our training recipe.

\begin{figure}[t]
    \centering
    % \vspace{-8pt} 
    \includegraphics[width=\linewidth]{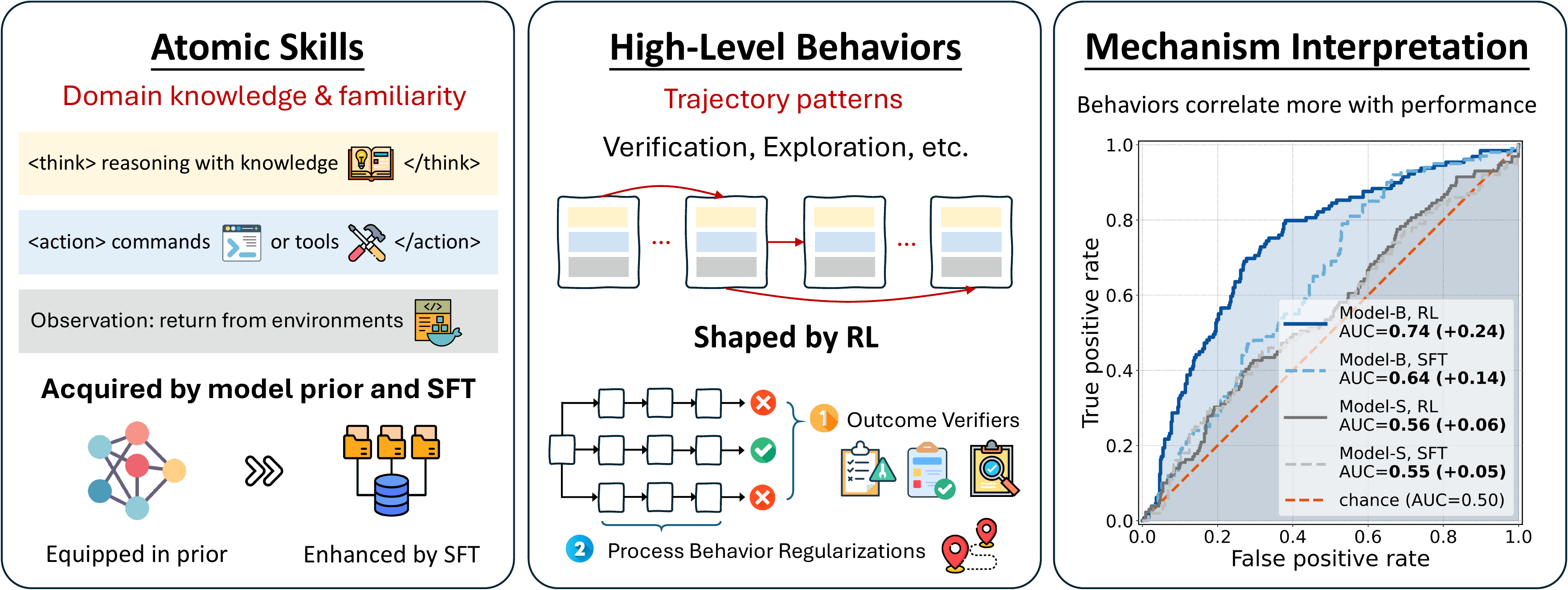}
    % \caption{
    % (\textbf{Left}) \textcolor{red}{TODO: explain agentic compositional generalization} Domain skills: domain knowledge and familiarity with specific tools and commands in single-turn actions are acquired through pretraining and SFT; (\textbf{Middle}) Multi-turn behaviors, which describe patterns of inter-turn actions, are shaped during RL; (\textbf{Right}) AUC-ROC results showing that (1) behaviors correlate more strongly with task performance, and (2) RL shapes behaviors. Details of the AUC-ROC tests are provided in Sec.~\ref{subsection:auc-roc-support}. } 
    % % \vspace
    \caption{
    The \textbf{Agentic Compositional Generalization} hypothesis. We consider terminal-agent capability at two levels that are acquired at different training stages. 
    (\textbf{Left}) Low-level skills, i.e., domain knowledge and familiarity with specific tools and commands that determine how each single-turn action is executed, are acquired during pretraining and SFT. (\textbf{Middle}) High-level behaviors, i.e., cross-turn patterns such as exploration and verification that determine what to do next, are shaped during RL by feedback from environment verifiers. Under our hypothesis, generalizability arises when RL-shaped behaviors compose and route the skills that an unseen task demands. (\textbf{Right}) Evidence consistent with this account: behavior features of a trajectory predict task success (AUC-ROC $0.74$), while the presence of individual skills is near chance (about $0.55$), and the link between behaviors and success strengthens from SFT to RL ($0.64$ to $0.74$); test details in Sec.~\ref{subsection:auc-roc-support}.
    }
    \label{fig:teaser-figure-triple}
\end{figure}

\subsection{Agentic Compositional Generalization}
\label{subsection:agentic-compositional-reasoning}

We model terminal-agent capability as a hierarchical decision-making process with two complementary levels of competence: \textbf{(i)} high-level \emph{behaviors} that determine \emph{what} the agent should do next (e.g., decomposing a task, planning future actions, or verifying current results), and \textbf{(ii)} low-level \emph{skills} that determine \emph{how} each action is executed (e.g., applying domain knowledge for reasoning or invoking appropriate tools). We formalize this process as a Markov Decision Process (MDP)~\citep{puterman2014markov}. At interaction step \(t\), the state \(s_t\) consists of the complete interaction history up to that point, including the user instruction, executed commands, and terminal feedback. An action \(a_t\) corresponds to the agent's single-turn output, including its reasoning tokens together with any tool invocation or terminal command. The policy \(\pi(a_t \mid s_t)\) defines the agent's action distribution. Within this formulation, we distinguish two complementary roles (see Fig.~\ref{fig:teaser-figure-triple}):

\begin{itemize}[leftmargin=*]

    \item \textbf{Low-level Single-turn Skills.} A skill is the ability to execute a single turn correctly given the current state $s_t$: applying domain knowledge, reasoning accurately, and issuing the appropriate command or tool call (e.g., composing a valid grep pipeline, running pytest, installing a package, see Appendix~\ref{appendix-sub:sft-data-coverage}). Skills are properties of individual actions, whether a step is locally correct, irrespective of the surrounding strategy. SFT primarily improves skills by concentrating the policy on valid, domain-appropriate actions (see Sec.~\ref{subsection:sft-skill-rl-behavior-support}).
    
    \item \textbf{High-level Behaviors.} A behavior is a pattern, typically spanning multiple turns, in how actions are chosen across turns, e.g., inspecting the environment before acting, verifying results before declaring completion, or abandoning a failing approach instead of repeating it (see App.~\ref{app:behavior-analysis-details}). Behaviors are properties of action sequences rather than of any single action: they manifest in multi-turn fragments and aggregate into trajectory-level patterns, so a trajectory can be flawed even when every step is locally correct. Our hypothesis is that behaviors compose, route, and coordinate low-level skills to accomplish long-horizon tasks; RL primarily shapes them (Sec.~\ref{subsection:auc-roc-support}).
\end{itemize}

%Examples and visualizations of these two levels are shown in 

%We model agentic terminal capability as a structured generation process that combines two levels of capability: (i) making a high-level plan, e.g., retrieving relevant information or planning future actions; and (ii) applying local operations, such as using domain knowledge for reasoning or making a specific tool call. To characterize these two roles formally, we cast terminal-agent operation as a Markov Decision Process (MDP)~\citep{puterman2014markov}, where the state $s_t$ is the interaction history up to turn $t$ (the instruction together with executed commands and their terminal feedback), an action $a_t$ is the single-turn operation the agent emits (a tool call or command, with its reasoning tokens), and the policy $\pi(a_t \mid s_t)$ is the model. On this basis, we define the two roles as: (a) low-level single-turn \textbf{Skills}: the quality of the per-step distribution $\pi(a_t \mid s_t)$, i.e., emitting a locally correct action from domain knowledge in the reasoning tokens and familiarity with the relevant tools; SFT primarily improves skills by concentrating $\pi(\cdot \mid s_t)$ on the valid, domain-appropriate subset of the action space; and (b) high-level multi-turn \textbf{Behaviors}: the sequential patterns over state-action pairs $(s_t, a_t)$ that the policy induces across a trajectory, such as routing low-level skills to solve the target problem. Examples and visualizations are provided in Fig.~\ref{fig:teaser-figure-triple}.

We propose the \textbf{Agentic Compositional Generalization} hypothesis, extending recent theories of compositional generalization for single-turn reasoning models~\citep{kong2026reasoning,yuan2026f} to interactive terminal agents. The main idea is that generalizability arises from the complementary roles of SFT and RL: pre-training and SFT equip the model with domain-specific low-level skills, while RL shapes high-level behaviors from environment feedback that compose, route, and coordinate these skills to solve long-horizon tasks. During deployment, the agent generalizes by combining shaped behaviors with the domain skills required by the target task. In this section, we present the intuition and implications of this hypothesis, while deferring the empirical validation to Sec.~\ref{section: analysis}, where we provide comprehensive experimental evidence supporting these claims.

%In this work, we propose the Agentic Compositional Generalization hypothesis, which is an extension of Compositional Generalization~\citep{kong2026reasoning,yuan2026f} for single-turn reasoning models, for terminal agents. This explains how OOD generalization arises in terminal-agent training: through pre-training and RL, the model acquires substantial low-level skills, and during RL it learns high-level behaviors from environment feedback. In evaluation, the model combines its shaped high-level behaviors with the low-level skills corresponding to the target domain. For the fluency of the presentation, we defer the experiments supporting these claims to Sec.~\ref{section: analysis} and providing hyperlinks and references where necessary. Here we primarily present the key findings and discuss how to design a training recipe based on them.

\textbf{Roles of SFT.}
Whether RL should be warm-started from an SFT model has been actively debated since DeepSeek-R1~\citep{guo2025deepseek}. Our findings suggest that the answer depends on the model's existing capabilities. For weaker base models, large-scale SFT plays a critical role by familiarizing the model with the target domains and interaction harness, enabling it to produce trajectories correct enough to yield meaningful learning signals during RL. More importantly, SFT equips the model with the low-level skills required for downstream tasks, providing the foundation upon which RL learns high-level behaviors that, under our hypothesis, enable generalizability (Sec.~\ref{subsection:sft-skill-rl-behavior-support} for experiments). In contrast, for stronger foundation models that already possess broad low-level skills and can adapt to new harnesses via their robust instruction-following capacity, we can perform zero-RL without SFT.

%\textbf{Roles of SFT.}
%Whether warm-starting RL with SFT is beneficial has been debated since DeepSeek-R1~\citep{guo2025deepseek}. In this work, we find that the answer depends on the base model's capabilities. For weak models, large-scale SFT is necessary to familiarize the model with the targeted domains and harness, so that it can obtain correct trajectories and thereby receive learning signals during RL; more importantly, the model gains or activates low-level skills, which lays the foundation for OOD generalization after the subsequent RL. Thus, initializing them with diverse low-level skills is necessary (see Sec.~\ref{subsection:sft-skill-rl-behavior-support}). For stronger models that already have good coverage of low-level skills and can adapt to the target harness through strong instruction-following, the SFT step before RL is optional.

\textbf{Roles of RL.}
Our findings are consistent with RL primarily reshaping \emph{high-level behaviors} while largely preserving the exhibited low-level skill repertoire. Across evaluation trajectories, the skills exhibited before RL remain strongly related to those exhibited after RL, while the task-specific combinations in which these skills appear change substantially (Sec.~\ref{subsection:sft-skill-rl-behavior-support} for experiments).
We hypothesize that this behavioral reorganization is an important driver of generalizability.
Consistent with this account, an AUC-ROC analysis~\citep{fawcett2006introduction} on evaluation trajectories shows that behavior-oriented trajectory features are substantially more predictive of task success than the presence of individual low-level skills (Fig.~\ref{fig:teaser-figure-triple} and Sec.~\ref{subsection:auc-roc-support}). Together, these observations support the hypothesis that an important role of agentic RL is to reshape reusable interaction behaviors and how previously exhibited skills are recruited and composed across tasks and domains.

%\textbf{Roles of RL.}
%During RL, the model preserves the low-level skills obtained from pre-training and SFT (see Sec.~\ref{subsection:sft-skill-rl-behavior-support}). Rather than gaining novel low-level skills, RL learns and shapes the high-level behaviors, which are closely related to OOD generalization performance (see Sec.~\ref{subsection:sft-skill-rl-behavior-support}). In addition, from an AUC-ROC test~\citep{fawcett2006introduction} on evaluation traces, we find that high-level behaviors correlate strongly with task performance, whereas the presence of specific skills shows little correlation with it (see Fig.~\ref{fig:teaser-figure-triple} and Sec.~\ref{subsection:auc-roc-support}). Consequently, when performing agentic RL, a key factor is enabling the model to learn generalizable behaviors through this process for better OOD performance.

% \subsection{How to shape behaviors efficiently?}

\begin{figure}[t]
    \centering
    % \vspace{-8pt} 
    \includegraphics[width=\linewidth]{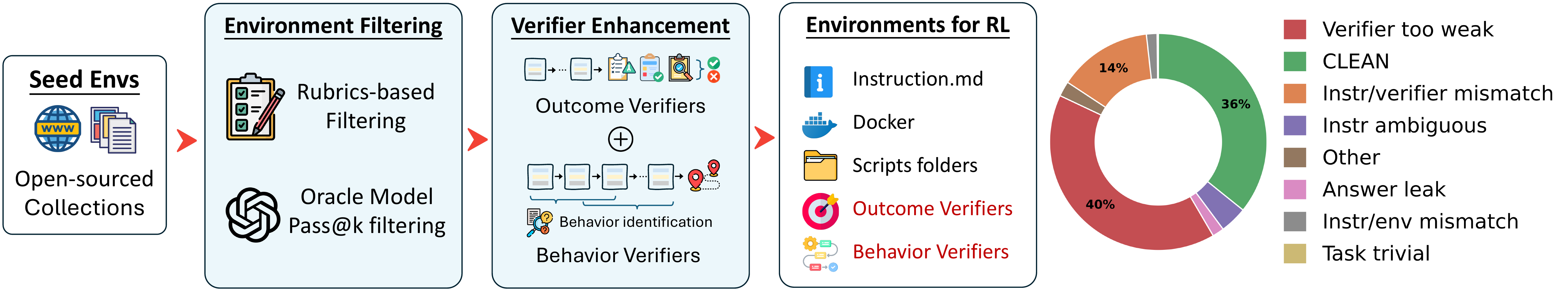}
    \caption{
    (\textbf{Left:}) Pipeline for environment filtering and verifier enhancement. (\textbf{Right:})  Environment quality evaluation results of TMax-15K~\citep{ivison2026tmax}.}
    % \vspace{-10pt}
    \label{fig:rl-pipeline}
\end{figure}

\subsection{\method: \methodlong}
\label{subsction:method-main}
% This work focus on agentic RL with synthetic terminal environments.

A synthetic terminal environment typically consists of an interactive sandbox, task-specific file systems, and a verifier that provides reward signals to the RL agent (Fig.~\ref{fig:rl-pipeline}). The discussion above suggests that effective terminal-agent RL training requires two complementary ingredients: \textbf{(1)} the starting checkpoint should have broad coverage of low-level atomic skills, and \textbf{(2)} the RL training environments should provide high-quality feedback that fosters the learning of reusable, high-level behaviors. To satisfy the first requirement, we construct a filtered version of the OpenThoughts~\citep{raoof2026openthoughts} SFT corpus that provides broad coverage of atomic skills for weaker models lacking strong prior capabilities (see App.~\ref{app:sft-training} for details and Sec.~\ref{subsection:sft-analysis} for ablations). To address the second, we propose \method, motivated by our observation that verifier quality is a key factor beyond environment quantity and diversity. Inaccurate verifiers can produce false-positive and false-negative reward signals on rollout trajectories, thereby inducing reward hacking (see App.~\ref{app:reward-hacking} for examples). Accordingly, \method\ improves the integrity of the reward signals provided by the source environment collection TMax~\citep{ivison2026tmax} through two key components: \textit{environment filtering} and \textit{verifier enhancement}.

%A synthetic terminal environment typically contains an interactive sandbox, file folders, and verifiers that provide signals to the RL agent, as shown in Fig.~\ref{fig:rl-pipeline}. From the analysis above, achieving good RL performance requires (1) Good starting models with high coverage on atomic skills; (2) high-quality environments fostering generalizable behaviors during the training process. For SFT, we prepare a filtered version of Openthoughts~\citep{raoof2026openthoughts} SFT collection, which has a high coverage on atomic skills for weak models who do not have good priors (See App.~\ref{app:sft-training} for details). For RL environments, our \method\ highlight the importance of verifiers, as they provide the signals that shape agent behaviors. \method\ has two key components:

\begin{wrapfigure}{r}{0.47\linewidth}
\vspace{-\intextsep}
\begin{minipage}{\linewidth}
\begin{algorithm}[H]
\small
\DontPrintSemicolon
\SetKwComment{Comment}{\textcolor{gray}{// }}{}
(1) SFT dataset construction with broad skill coverage and SFT training
\textcolor{gray}{\# Details in App.~\ref{app:sft-training}, Discussions in Sec.~\ref{subsection:sft-analysis}} \\
% $^\dagger$
\colorbox{blue!8}{\parbox{\dimexpr\hsize-2\fboxsep\relax}{%
\textcolor{blue}{\# \method\ core components:}\\
(2) RL Environment Filtering \\
(2.1) Deploy Rubrics-based Filtering; \\
(2.2) Oracle Agent Pass@$k$ Filtering; \\
(3) Verifier Enhancement (\ref{eq:shaped-adv})%
}}\\
(4) RL on final envs\;
% \vspace{2pt}
% {\footnotesize $^\dagger$\textcolor{gray}{optional; only for our 8B weak models.}}
\caption{Full Training recipe.}
\label{alg:recipe}
\end{algorithm}
\end{minipage}
\vspace{-\intextsep}
\end{wrapfigure}

\textbf{Environment filtering.}
Although synthetic environments have become a dominant paradigm for terminal-agent RL, their quality remains highly variable. Existing open-source collections often contain weak or mismatched verifiers, resulting in the potential for noisy supervision that encourages shortcut solutions, reward hacking, or overfitting. Since verifier quality directly shapes the high-level behaviors learned during RL, ensuring environment quality is a prerequisite for effective training. We develop a lightweight environment filtering system that evaluates synthetic environments using a set of rubrics without executing the underlying sandbox. Our framework identifies eight categories of quality issues, including weak verifiers, mismatched task specifications, and other inconsistencies (see App.~\ref{app:env-audit} for details and examples). Applying this evaluation to the TMax~\citep{ivison2026tmax} environment collection reveals that more than $60\%$ of the environments exhibit at least one quality issue, leaving fewer than $40\%$ suitable for RL training\footnote{Our filtering process may produce both false positives and false negatives. Also, there might be multiple issues in one environment. Details of environment filtering can be found in App.~\ref{app:env-audit}. Examples of reward hacking with unfiltered environments can be found in App.~\ref{app:reward-hacking}.} (see Fig.~\ref{fig:rl-pipeline}).

We further filter the remaining environments using GPT-5.4~\citep{openai2026gpt54card} with harness of Terminus-2~\citep{harbor2026} as an oracle verifier. Specifically, we perform a pass@$k=2$ evaluation and remove environments with pass@$k=0$ 
% \semih{Arm-(2) of my proposed ablation above would quantify this}
, which are either excessively difficult or incompatible with our training infrastructure. The resulting curated collection, \textbf{\method-TMax-3.5K}, serves as the training environment set used throughout this work.

\textbf{Verifier enhancement.}
Environment filtering improves the quality of training tasks, but each environment still provides only sparse outcome supervision. We introduce an optional \emph{verifier enhancement} step that supplies denser behavior-level feedback during RL. The key idea is to complement the trajectory-level reward with lightweight behavior verifiers that identify desirable or undesirable interaction patterns and provide turn-level shaping signals. These signals encourage generalizable behaviors while remaining compatible with standard RL objectives.

We instantiate this idea using GRPO~\citep{shao2024deepseekmath}. Unless otherwise stated, we use a binary outcome reward \(r_i\in\{0,1\}\): for each prompt, we sample a group of \(G\) rollouts, assign \(r_i=1\) only if the $i$-th rollout passes all verifiers and test cases, normalize rewards within the group, and broadcast the resulting scalar advantage uniformly to every response token. While effective, this trajectory-level signal is blind to behaviors that occur during interaction. To incorporate behavior supervision, we augment the standard GRPO advantage \(A_{i,t}\) with a turn-wise shaping term \(s_{i,t}\):
\begin{equation}
\label{eq:shaped-adv}
A_{i,t}
  =
  \frac{r_i-\bar r}{\sigma_r+\epsilon},
  \qquad
  \bar r=\frac{1}{G}\sum_{j=1}^{G}r_j,
  \qquad
  \sigma_r=\mathrm{std}(\{r_j\}_{j=1}^{G}),
  \qquad
  \tilde A_{i,t} = A_{i,t}+s_{i,t}
\end{equation}
% \begin{equation}
% \tilde A_{i,t}
% =
% A_{i,t}+s_{i,t},
% \label{eq:shaped-adv}
% \end{equation}
where \(\bar r\) and \(\sigma_r\) denote the within-group reward mean and standard deviation, and \(\epsilon\) is a small constant. The shaping term \(s_{i,t}\), applied after reward normalization, may be positive to reinforce desirable behaviors or negative to discourage undesirable ones.
One particularly effective \(s_{i,t}\) uses a simple rule-based verifier that penalizes repetitive interaction loops with little information gain. Specifically,
\[
s_{i,t}
=
-\delta\,
\mathbf 1
\!\left[
t\in
\bigcup_{k\in\mathcal L_i}
\mathcal T(k)
\right],
\]
% where \(\delta=0.05\) shafiq{Need some note about why 0.05}, 
% where \(\delta=0.05\), 
where $\delta$ is a hyperparameter~\footnote{We used $0.05$ in our experiments as determined by a lightweight search.}, \(\mathcal T(k)\) denotes the response-token indices of turn \(k\), and \(\mathcal L_i\) is the set of turns identified as repetitive. 
We penalize repetition behaviors: 
a turn is flagged as repetitive when both its command and observation have Jaccard similarity greater than $0.8$ with an earlier turn. We also provide experiments with additional behavior verifiers in Sec.~\ref{subsection:exp-qwen3}. \

%\textbf{Verifier enhancement.} In addition to filtering, we provide an optional verifier enhancement step that supplies denser signals to shape behaviors during RL. In practice, we apply turn-wise advantage shaping in the GRPO objective based on rule-based identified behavior patterns. 
% \textbf{Behavior-level advantage shaping.}

%We train with GRPO~\citep{shao2024deepseekmath} under a sparse binary outcome reward $r_i\in\{0,1\}$ unless otherwise stated: for each prompt we sample a group of $G$ rollouts, each one receive $r=1$ only if it can pass all verifiers/test cases, standardize their rewards within the group, and broadcast the scalar advantage to every response token of the rollout. This trajectory-level signal is uniform over a rollout and blind to \emph{local} behaviors. To shape behaviors with behavior verifiers, we add a turn-wise shaping term $s_{i,t}$ on top of the GRPO advantage ${A}_{i,t}$:
%\begin{equation}
%{A}_{i,t}
%  \;=\;
%  \big(r_i - \bar{r}\big)\big/\big(\sigma_r + \epsilon\big), \ 
  \ 
%  \bar{r}=\sum\nolimits_{j=1}^{G} r_j / G,
%  \ 
%  \sigma_r=\mathrm{std}(\{r_j\}_{j=1}^{G}), \quad
%  \tilde{A}_{i,t}
%  \;=\;
%  {A}_{i,t}
%  \;\;{\color{red}+\;\; s_{i,t}}
%  \label{eq:shaped-adv}
%\end{equation}
%where $\bar{r}$ and $\sigma_r$ are the within-group reward mean and standard deviation and
%$\epsilon$ a small constant. The highlighted {\color{red}$s_{i,t}$}, applied after
%normalization, can be \emph{negative} to penalize an undesirable
%behavior or \emph{positive} to reward a desirable one. 

%In the main experiments, we penalize repetitive loops with low information gain. More details and results for other behavior verifiers are provided in Sec~\ref{subsection:exp-qwen3}.
%For our repetitive-loop penalty we set
%$s_{i,t}=-\,\delta\,\mathbf{1}\!\left[\, t \in \bigcup_{k \in \mathcal{L}_i} \mathcal{T}(k) \,\right]$
%with $\delta=0.05$, where $\mathcal{T}(k)$ are the response-token indices of turn $k$ and
%$\mathcal{L}_i$ is a rule-based set of loop turns, flagged when a turn's command and
%observation' Jaccard-similarity  both exceed threshold $\kappa=0.8$ against an earlier turn.

\textbf{Recipe summary and intuition.}
Our overall training recipe is summarized in Alg.~\ref{alg:recipe}. For weaker base models, we include an SFT stage to establish broad coverage of low-level skills before RL. Additional implementation details, including the agent harness, datasets, and training configurations, are provided in Sec.~\ref{section:experiments-main} and App.~\ref{app:training-details}. The central intuition is simple: once a model possesses sufficient low-level skills, RL should focus on learning reusable high-level behaviors. Providing accurate verifier signals through high-quality environments enables these behaviors to generalize across domains, leading to improved out-of-distribution performance.

%\textbf{Recipe summary and intuitions.}
%Our training recipe is summarized in Alg.~\ref{alg:recipe}: for weaker models, we add an SFT stage to provide the necessary low-level skills. More training configurations, such as harness and data details, are available in Sec.~\ref{section:experiments-main} and App.~\ref{app:training-details}. The intuition is that providing accurate signals to models that already have sufficient coverage of the necessary low-level skills fosters generalizable high-level behaviors and leads to OOD generalization.

\section{Experiment Results}
\label{section:experiments-main}
In this section, we present the evaluation results of our \method\ pipeline with two base model families: Qwen3 and Qwen3.5~\footnote{Qwen3 uses full attention with GQA, while Qwen3.5 uses a hybrid architecture with Gated DeltaNet linear attention and full attention blocks. For the 27B experiment, we use Qwen3.6, following~\cite{ivison2026tmax}, which has the same architecture as Qwen3.5.}~\citep{yang2025qwen3, team2026qwen35, team2026qwen36}. We use four terminal benchmarks for evaluation: \textbf{(OpenThoughts-)Terminal-Bench-Lite}~\citep{raoof2026openthoughts}, \textbf{Terminal-Bench-v2.1}~\citep{merrill2026terminal}, \textbf{Terminal-World-Verified}~\citep{chu2026terminalworld}, and \textbf{Terminal-Bench-Pro}~\citep{wang2025letflowagenticcrafting}. For brevity, we refer to them as \texttt{otbl}, \texttt{tbv2.1}, \texttt{twv}, and \texttt{tbpro}, respectively, when there is no ambiguity. We also hold out a subset of around $300$ tasks from the TMax collection, excluding them from training and using them for controlled experiments. We adopt the default evaluation criteria of each benchmark: \texttt{otbl} uses a continuous score in $[0, 1]$, while the others use a binary score, assigning $1$ only if the solution passes all verifiers (unit tests).

\subsection{Qwen3-8B Model Family}
\label{subsection:exp-qwen3}
\textbf{Training.} Using \texttt{Qwen3-8B} as the base model, we apply the full \method\ pipeline. We filter out trajectories from the OpenThoughts-100K SFT collection~\citep{raoof2026openthoughts} that cannot be converted from its original Terminus-2 harness format to our Endless-Agent harness, resulting in $78$K trajectories. SFT is done with LLaMA-Factory~\citep{zheng2024llamafactory}. For RL, we adopt Endless-Agent as the agent harness~\citep{gandhi2026endless}. The RL training pipeline is also based on Endless-Agent, which in turn builds on SkyRL~\citep{cao2025skyrl}, Harbor~\citep{harbor2026}, and Verl~\citep{sheng2025hybridflow}. We use the \textbf{\method-TMax-3.5K} environment set as our RL training set.

\textbf{Baselines and Evaluation.} We compare our method with open-source fine-tuned $8$B models as well as the base model. The baselines include \href{https://huggingface.co/open-thoughts/OpenThinkerAgent-8B-RL}{OpenThinker-8B-RL} and \href{https://huggingface.co/open-thoughts/OpenThinker-Agent-v1}{OpenThinker-Agent-v1}~\citep{guha2026openthoughts, raoof2026openthoughts}, both of which perform SFT before RL on their synthetic environments; \href{https://huggingface.co/obiwan96/qwen3-8b-openthinker-sft-endless-terminals}{OT-SFT-Endless-8B}, which performs RL on the Endless-Terminals collection starting from the OpenThoughts SFT checkpoint~\citep{gandhi2026endless}; and \href{https://huggingface.co/allenai/tmax-8b}{TMax-8B} and \href{https://huggingface.co/allenai/tmax-sft-8b}{TMax-SFT-8B}~\citep{ivison2026tmax}, which are trained using their synthetic RL environments and SFT data, respectively. Each baseline is evaluated using its corresponding harness. We set the interaction turn budget to $64$ and keep all other evaluation configurations fair across methods. Further evaluation details are provided in App.~\ref{app:evaluation}.

\paragraph{Main Results.}
\newcommand{\gpm}[2]{#1\,{\scriptsize$\pm$#2}}
\newcommand{\gpmb}[2]{\textbf{#1}\,{\scriptsize$\pm$#2}}
\begin{table}[t]
\centering
\small
\setlength{\tabcolsep}{5pt}
\resizebox{1.\linewidth}{!}{
\begin{tabular}{lllccccc}
\toprule
Model & Harness & Training & Otbl & Tbv2.1 & Tbpro & Twv & \textbf{Avg.} \\
\midrule
River-8B (ours) & EndlessAgent & SFT+RL & \gpmb{21.8}{2.9} & \gpmb{9.7}{0.6} & \gpmb{23.0}{3.1} & \gpmb{23.0}{2.5} & \gpmb{19.4}{1.2} \\
OpenThinker-8B-RL & Terminus-2 & SFT+RL & \gpm{18.4}{1.5} & \gpm{8.6}{0.6} & \gpm{21.8}{1.6} & \gpm{22.3}{1.5} & \gpm{17.8}{0.7}  \\
OT-SFT-Endless-8B & EndlessAgent & SFT+RL & \gpm{14.1}{3.6} & \gpm{6.4}{1.7} & \gpm{16.8}{0.8} & \gpm{21.7}{1.9} & \gpm{14.7}{1.1} \\
TMax-RL-8B & Vanillux2 & RL & \gpm{17.0}{3.7} & \gpm{6.0}{0.6} & \gpm{16.3}{1.0} & \gpm{19.0}{2.6} & \gpm{14.6}{1.2} \\
OpenThinker-Agent-v1 & Terminus-2 & SFT+RL & \gpm{16.0}{0.6} & \gpm{5.2}{0.6} & \gpm{16.2}{1.0} & \gpm{21.7}{2.9} & \gpm{14.8}{0.8} \\
\midrule
TMax-SFT-8B & Vanillux2 & SFT & \gpm{12.1}{3.4} & \gpm{4.5}{1.9} & \gpm{13.7}{1.9} & \gpm{14.3}{4.1} & \gpm{11.2}{1.5} \\
River-SFT-8B & EndlessAgent & SFT & \gpm{11.3}{2.3} & \gpm{6.4}{1.3} & \gpm{13.7}{2.6} & \gpm{18.5}{2.3} & \gpm{12.5}{1.1} \\
\midrule
Qwen3-8B & EndlessAgent & None & \gpm{3.8}{1.3} & \gpm{2.2}{1.2} & \gpm{6.8}{0.6} & \gpm{9.8}{1.6} & \gpm{5.7}{0.6} \\
\bottomrule
\end{tabular}
}
\caption{Evaluation results of $8$B agents on four terminal benchmarks. Numbers are benchmark scores ($\times10^{2}$). \textbf{Avg.}\ is the unweighted mean of the four benchmarks. Scores are mean\,$\pm$\,std over $3$ seeds. Baseline scores are reported by our evaluation pipeline under fair comparison.}
\label{tab:leaderboard_main}
\end{table}

Comparisons are summarized in Tab.~\ref{tab:leaderboard_main}. We observe that our model achieves the best overall performance. Among the baselines, OT-SFT-Endless-8B uses the same harness and also starts from SFT data from the OpenThoughts collection, but with lower atomic-skill coverage. Its RL stage is trained on a synthetic environment set of $3.3$K tasks, which is comparable in scale to our collection. With the harness and RL training pipeline held fixed, the main differences in this comparison are the SFT initialization/data and the RL environment set; we examine these factors separately through controlled experiments in Sec.~\ref{section: analysis}. 
TMax-8B uses a different harness, vanillux2, a mini-swe-agent-2~\citep{yang2024sweagent} variant, and is trained on the full TMax-15K environment collection. Despite using substantially fewer RL environments, our model achieves consistently higher performance across all four evaluation benchmarks. As discussed in App.~\ref{app:reward-hacking}, the unfiltered TMax collection also contains environments with verifier failures that can produce misleading reward signals and reward-hacking opportunities.
The OpenThinker models use the Terminus-2 harness~\citep{harbor2026} and are trained with large-scale SFT and RL. OpenThinker-8B-RL achieves the strongest performance among the baselines, while our model still maintains a clear average performance advantage. Possible contributors to this gap include broader SFT skill coverage, higher-quality RL verifier signals from environment filtering, and behavior-level regularization, which we study separately in Sec.~\ref{section: analysis}.

% \input{table_folder/main/AUC_ROC_small}

% \textbf{Behavior Ablations and Analysis.} Regarding verifier enhancement introduced  in Sec.~\ref{subsction:method-main}, we consider the repetition behavior as a turn-wise penalty. This behavior is selected because it ranks \#2 in the single-behavior-feature AUC-ROC test, indicating a strong correlation with task performance. The rank-\#1 behavior is \texttt{verify\_before\_done}. As shown in Tab.~\ref{tab:behavior-auc-compact}, we also run experiments that provide an additional turn-wise reward based on it. The performance and behavior comparisons for both behaviors are visualized in Fig.~\ref{fig:main-behavior-ablation}. Compared with GRPO using ORM only, we find that advantage shaping does reshape the target behaviors: it reduces the repetition ratio and increases the fraction of verification in the corresponding runs. However, only the repetition-penalty run yields a noticeable overall performance improvement, whereas rewarding verification brings no additional gains. We interpret this as indicating that verification is correlated with task performance but does not directly cause task success.

\begin{figure}[t]
    \centering
    % \vspace{-8pt} 
    \includegraphics[width=\linewidth]{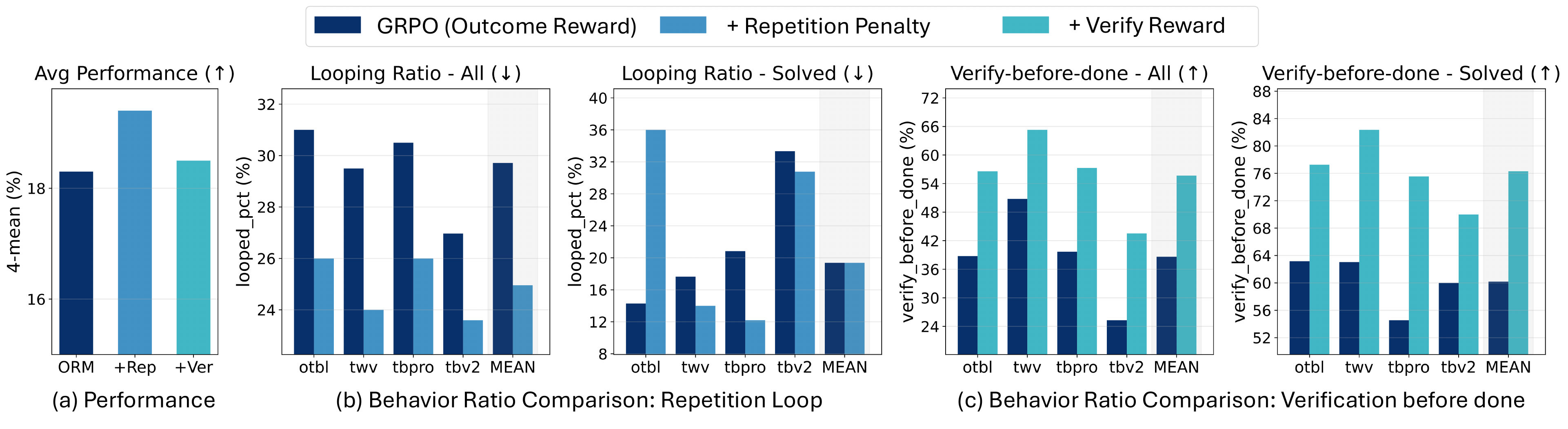}
    \caption{
    Ablations on behavior verifiers. \textbf{(a)}: Averaged performance on the four evaluation benchmarks; \textbf{(b, c)}: Behavior ratio comparison between RL with and without behavior verifiers. Arrows $\uparrow/\downarrow$: the higher/lower, the better.}
    % \vspace{-20pt}
    \label{fig:main-behavior-ablation}
\end{figure}

\textbf{Behavior Ablations and Analysis.}
We select behavior verifiers for the verifier enhancement step in a data-driven way. We first assess the utility of each candidate behavior by how well it predicts the task success, using the single-behavior-feature AUC-ROC test (see Sec.~\ref{subsection:auc-roc-support} for AUC-ROC details) shown in Tab.~\ref{tab:behavior-features} in App.~\ref{app:behavior-analysis-details}. The two strongest signals are \texttt{verify\_before\_done} and \texttt{repetition\_rate}. We then run experiments with each of these top behaviors as a turn-level shaping signal, with the sign chosen according to its correlation: a turn-wise reward for verification before completion, and a turn-wise penalty for repetitive loops. The performance and behavior comparisons for both behaviors are visualized in Fig.~\ref{fig:main-behavior-ablation}. Compared with outcome-based reward only, we find that advantage shaping does reshape the target behaviors: it reduces the repetition ratio and increases the fraction of verification in the corresponding runs. However, only the repetition-penalty run yields a noticeable overall performance improvement, whereas rewarding verification brings no additional gains. We interpret this as indicating that verification is correlated with task performance but does not directly cause task success. Based on this result, only the repetition penalty enters the final RIVER recipe as its verifier enhancement component~\ref{subsction:method-main}.

We hope this discussion draws more attention to systematic study of multi-turn behaviors that are causally related to task performance, and to the design of advantage shaping in RL. Further ablations, such as the selection of SFT data and RL environments, are discussed in Sec.~\ref{subsection:sft-analysis} and~\ref{subsection:rl-analysis}.

\begin{AIbox}{Section~\ref{subsection:exp-qwen3} Takeaways}
$\bullet$ Our \method-8B model achieves the best performance on four terminal benchmarks.\\
% $\bullet$ Advantage shaping from behavior verifiers can further shape behaviors, and discovering more beneficial behavior verifiers is an interesting and non-trivial direction.
$\bullet$ Advantage shaping from behavior verifiers steers the targeted behaviors, and discovering behavior verifiers that improve performance is an interesting and non-trivial direction.

\end{AIbox}

\subsection{Qwen3.5 and Qwen3.6 Families}
\label{subsection: qwen3.5 and qwen3.6 experiment-main}
% A figure (Pareto Frontier) or a table. Making comparison with TMax Qwen3.5 series models. 
\textbf{Training.} For an apples-to-apples comparison, we keep the same base models as the TMax collection: Qwen3.5-\{2B, 4B, 9B\} and Qwen3.6-27B. We also adopt their training pipeline and most of their configurations, including the harness and their RL algorithm DPPO~\citep{qi2026rethinking}, except for a few changes made under resource constraints, such as switching from the sandbox API to locally hosted Docker containers. For evaluation, we use the two benchmarks reported in their paper. 
% Training details are available in the App.~\ref{app:training-details}. TODO: Adding experiment details
These experiments can be viewed as an ablation on environment quality: if we keep only the environments after our filtering process, can we further improve the trained model's performance?

\begin{figure}[b]
    \centering
    % \vspace{-8pt} 
    \includegraphics[width=\linewidth]{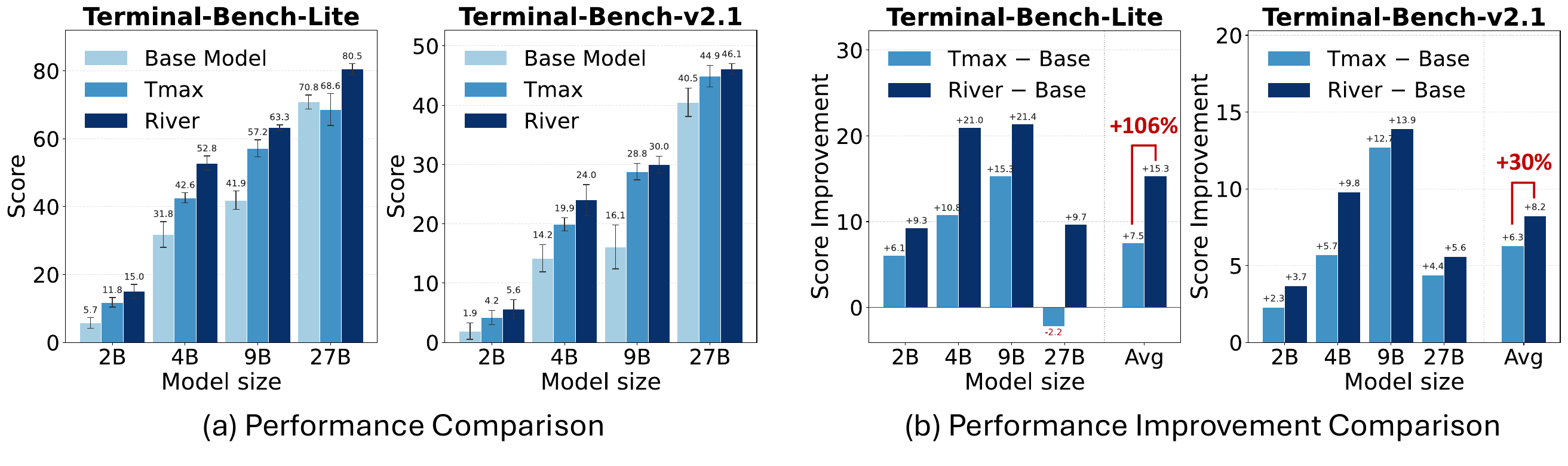}
    \caption{
    Comparison results with the TMax model series. (a) Performance comparison from 2B to 27B models; (b) RL performance improvement over base models from 2B to 27B models. We report the mean value and standard deviation with $3$ seeds. TMax results are adopted from their report.}
    % \vspace{-10pt}
    \label{fig:tmax-series-exp}
\end{figure}

\textbf{Main Results.} The evaluation results are presented in Fig.~\ref{fig:tmax-series-exp}. We observe that \method\ consistently improves base-model performance across model sizes and outperforms the original TMax on both Terminal-Bench-Lite and Terminal-Bench-v2.1. These results demonstrate that our method generalizes across model families, model sizes, and agent harnesses in agentic RL for terminal agents.
We also report the performance gains of the RL models over their corresponding base models.
Notably, using only $\sim\!30\%$ of the Tmax environments (our TMax-\method-3.5K set), we achieve, on average, $106\%$ and $30\%$ larger RL gains on Terminal-Bench-Lite and Terminal-Bench-v2.1, respectively.

\begin{AIbox}{Section~\ref{subsection: qwen3.5 and qwen3.6 experiment-main} Takeaways}
$\bullet$ With our filtered environment set, \textbf{\method-TMax-3.5K}, RL using the same base models achieves on average $106\%$ and $30\%$ larger performance gains on \texttt{otbl} and \texttt{tbv2.1}, respectively, compared with TMax training on the full environment set. \\
$\bullet$ The comparison between \method\ and TMax shows that \method\ generalizes across model families, model sizes, agent harnesses, and RL algorithms.
\end{AIbox}

\section{Analysis}
\label{section: analysis}
In this section, we first present experiments in Secs.~\ref{subsection:sft-skill-rl-behavior-support} and \ref{subsection:auc-roc-support} that support our \textbf{Agentic Compositional Generalization} hypothesis introduced in Sec.~\ref{subsection:agentic-compositional-reasoning}. We then provide further analyses and ablations of SFT in Sec.~\ref{subsection:sft-analysis} and RL in Sec.~\ref{subsection:rl-analysis}. Unless otherwise stated, we use Qwen3-8B as the base model.

\subsection{Skill preserving and behavior shaping in RL}
\label{subsection:sft-skill-rl-behavior-support}
We would like to understand how RL changes the model's capabilities over low-level skills and high-level behaviors. We compare evaluation traces before and after RL for our 8B models, which are warm-started with SFT. With a list of $N_s=60$ pre-defined low-level skills, we use GPT-5.4 to extract the feature of each evaluation trace as a binary vector indicating which skills are exhibited in that trace. The evaluation benchmarks contain $N_t=589$ tasks in total, so we obtain two binary matrices $M^{\mathrm{pre}}, M^{\mathrm{post}} \in \{0,1\}^{N_s\times N_t}$ (before and after RL), where $M_{s,t}=1$ iff skill $s\in\{1,\dots,N_s\}$ is exhibited in the trace for task $t\in\{1,\dots,N_t\}$. Row $M_{s,:}\in\{0,1\}^{N_t}$ is skill $s$'s \emph{task-incidence vector} (on which tasks it appears); column $M_{:,t}\in\{0,1\}^{N_s}$ is task $t$'s \emph{skill vector} (which skills it recruits).

\begin{figure}[t]
    \centering
    % \vspace{-8pt} 
    \includegraphics[width=\linewidth]{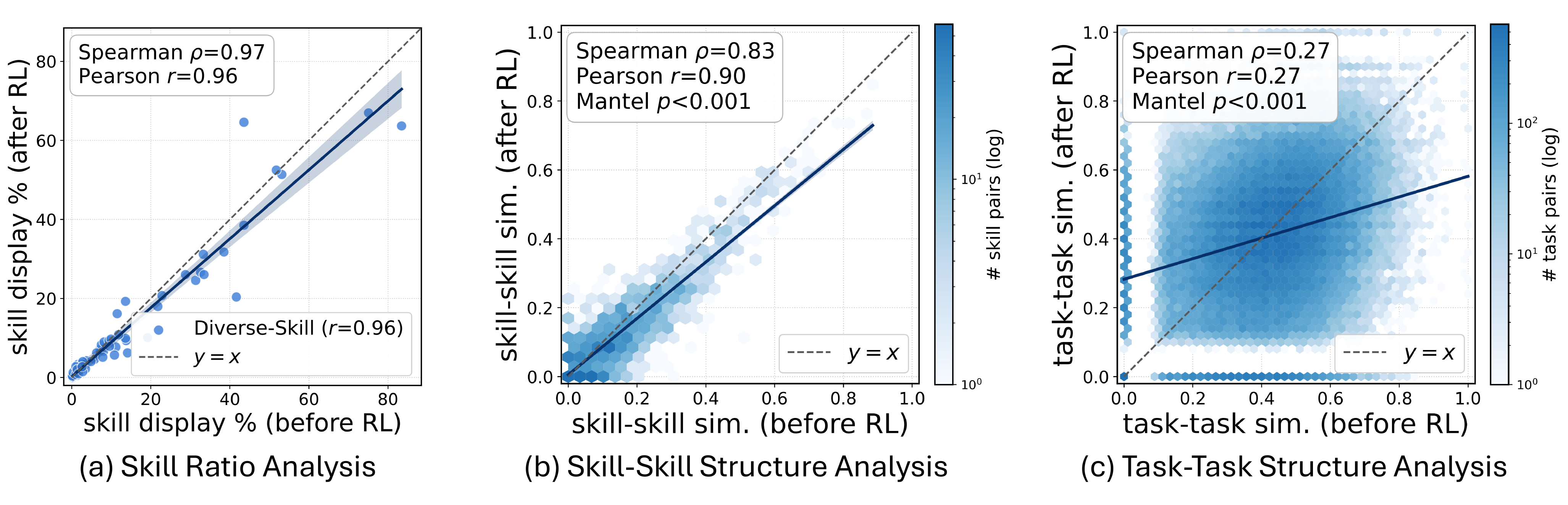}
    \caption{ Skills and behaviors correlation analysis before and after RL. (a) Correlation between skill ratio; (b) Correlation between skill-skill occurrence; (c) Correlation between task-task structure.
     }
    % \vspace{-10pt}
    \label{fig:mechanism-datasupport}
\end{figure}

First, we compute the exhibition ratio of each skill by averaging along the task dimension, $r_s = \tfrac{1}{N_t}\sum_{t=1}^{N_t} M_{s,t}$, once for $M^{\mathrm{pre}}$ and once for $M^{\mathrm{post}}$. The results are presented in Fig.~\ref{fig:mechanism-datasupport}~(a), where each point corresponds to one skill $s$ at coordinates $(r_s^{\mathrm{pre}}, r_s^{\mathrm{post}})$. We observe that the skill ratios remain strongly correlated: skills that are rarely exhibited before RL also tend to remain rarely exhibited afterward. This suggests that RL does not substantially expand the observed skill repertoire beyond that already expressed by the SFT checkpoint.

Second, in Fig.~\ref{fig:mechanism-datasupport}~(b and c) we apply Representational Similarity Analysis (RSA)~\citep{kriegeskorte2008representational} to the same matrices $M^{\mathrm{pre}}, M^{\mathrm{post}}$ along their two axes. The {skill-skill} structure is the matrix $S^{\mathrm{skill}}\in\mathbb{R}^{N_s\times N_s}$ with $S^{\mathrm{skill}}_{ij}=\cos\!\big(M_{i,:},\,M_{j,:}\big)$, the cosine similarity between every pair of skills over their task-incidence vectors, i.e., ``which skills co-occur''. Symmetrically, the {task-task} structure is $S^{\mathrm{task}}\in\mathbb{R}^{N_t\times N_t}$ with $S^{\mathrm{task}}_{ab}=\cos\!\big(M_{:,a},\,M_{:,b}\big)$, the cosine similarity between every pair of tasks over their skill vectors, i.e., ``which tasks are solved with the same skills''. We build each structure once for $M^{\mathrm{pre}}$ and once for $M^{\mathrm{post}}$; each panel is a density scatter of one structure's upper-triangle entries before RL ($x$) vs after RL ($y$), one point per pair. We report the Pearson~\citep{pearson1896vii} and Spearman correlation~\citep{spearman1961proof} between the two triangles, with a Mantel permutation test~\citep{mantel1967detection} for significance.
% Left = skill-skill ($\rho=0.83$, $p<0.001$); right = task-task ($\rho=0.27$, $p<0.001$).

Fig.~\ref{fig:mechanism-datasupport}~(b, c) support the hypothesis that 
(1) \textbf{The skill-skill structure is strongly preserved during RL}: in Fig.~\ref{fig:mechanism-datasupport}~(b), points lie close to $y=x$, with $\rho=0.83$, indicating that skills that co-occurred before RL tend to continue co-occurring afterward. In contrast, (2) \textbf{the task-task structure changes substantially}: in Fig.~\ref{fig:mechanism-datasupport}~(c), correlation is $\rho=0.27$, with a diffuse cloud far from $y=x$, suggesting that the set of skills recruited for a given task is reorganized during RL. Together, these results are consistent with RL primarily reshaping how existing skills are selected and composed across tasks.

\subsection{Behaviors' Correlation with Task Performance}
\label{subsection:auc-roc-support}
In this section, we provide additional details on the experimental design and results of the Area Under the Receiver Operating Characteristic Curve (\textbf{AUC-ROC}) analysis~\citep{fawcett2006introduction} introduced in Sec.~\ref{subsection:agentic-compositional-reasoning}. We test whether task success is associated more with which \emph{skills} a trajectory uses or with how the agent \emph{behaves}. For each evaluation trajectory, we construct two representations and use each to predict the binary outcome \texttt{solved} with an $\ell_2$-regularized logistic regression classifier, evaluated by ROC-AUC using $15$ seeds and $5$-fold cross-validation. \textbf{Model-S (skills)} encodes the skills that appear in the evaluation traces, together with the task categories, whereas \textbf{Model-B (behavior)} encodes agent behaviors using $12$ rule-based features, such as verify-before-done and repetition. We also perform human inspection to remove behaviors with trivial meanings. We compute both representations from the traces generated by the SFT and RL checkpoints (trained without behavior-verifier enhancement). ROC-AUC measures the probability that the classifier assigns a higher score to a randomly chosen solved task than to a randomly chosen unsolved task. Thus, $0.5$ corresponds to chance-level discrimination, while $1.0$ corresponds to perfect separation. An AUC near $0.5$ therefore indicates that the corresponding features provide little information about whether a task is solved. Further details of the setup are provided in App.~\ref{app:behavior-analysis-details}. 
% \textcolor{red}{(TODO)} \textcolor{red}{TODO: mentioning human selection on behaviors.}

Test results and visualization of curves are presented in Fig.~\ref{fig:teaser-figure-triple} (Right). We make two main observations from the results. (i) Behavior-oriented trajectory features are substantially more predictive of success than skill-presence features: Model-B reaches AUC = 0.64 to 0.74, while Model-S remains near chance (0.48 to 0.56). Knowing which skills appear in a trajectory therefore provides relatively little information about whether the task is solved, whereas how the agent behaves is substantially more predictive. (ii) This association strengthens after RL: from SFT to RL, the AUC of Model-B increases from 0.64 to 0.74, while that of Model-S remains nearly unchanged at approximately 0.55. Together with the structural analysis in Sec.~\ref{subsection:sft-skill-rl-behavior-support}, these results are consistent with a larger change along the trajectory-level behavior axis than in the exhibited skill repertoire, supporting our hypothesis that RL primarily reshapes how existing skills are used across turns and tasks.

\begin{AIbox}{Section~\ref{subsection:sft-skill-rl-behavior-support} and~\ref{subsection:auc-roc-support} Takeaways}
$\bullet$ RL largely preserves the exhibited low-level skill repertoire while reorganizing how skills are recruited across tasks. \\
$\bullet$ Task success is associated more strongly with trajectory-level behavioral features than with which skills a trajectory exhibits. \\
$\bullet$ These results are consistent with RL reshaping high-level behaviors that route and compose previously exhibited single-turn skills.
\end{AIbox}

% \begin{itemize}
%     \item RL largely preserves the exhibited low-level skill repertoire while reorganizing how skills are recruited across tasks.
%     \item Task success is associated more strongly with trajectory-level behavioral features than with which skills a trajectory exhibits.
%     \item These results are consistent with RL reshaping high-level behaviors that route and compose previously exhibited single-turn skills.
% \end{itemize}

\subsection{SFT Analysis}
\label{subsection:sft-analysis}
This section provides additional experimental results and evidence supporting the importance of performing SFT for broader atomic-skill coverage. We use Qwen3-8B and a public open-source SFT dataset collection from the OpenThoughts~\citep{guha2026openthoughts, raoof2026openthoughts}. We construct three different sets of SFT data after our format converting and filtering process: (1) \textbf{Narrow-SFT}, with narrow skill coverage: the data comes mainly from \texttt{nl2bash} and \texttt{InferredBugs} and contains $\sim\!13.6$K trajectories; (2) \textbf{Diverse-SFT}, with diverse coverage of atomic skills: the data is collected from a wider range of sources, including \texttt{SWE-Smith}~\citep{yang2026swe} and Openthoughts' own IssueTasks datasets, and contains $\sim\!78$K trajectories after filtering; and (3) \textbf{Diverse-DS-SFT}, a random subset version of \textbf{Diverse-SFT} containing $13.6$K trajectories (similar size as Narrow-SFT). 
A visualization of skill coverage is presented in Fig.~\ref{fig:atomic-coverage-3way} (in App.~\ref{appendix-sub:sft-data-coverage}), where we observe that the \texttt{Narrow} set covers only a subset of the atomic skills, whereas the \texttt{Diverse} set achieves higher coverage. We also add a zero-RL group, in which RL is performed directly without SFT.

\begin{wrapfigure}{r}{0.4\linewidth}
    \centering
    \vspace{-8pt}
    \includegraphics[width=\linewidth]{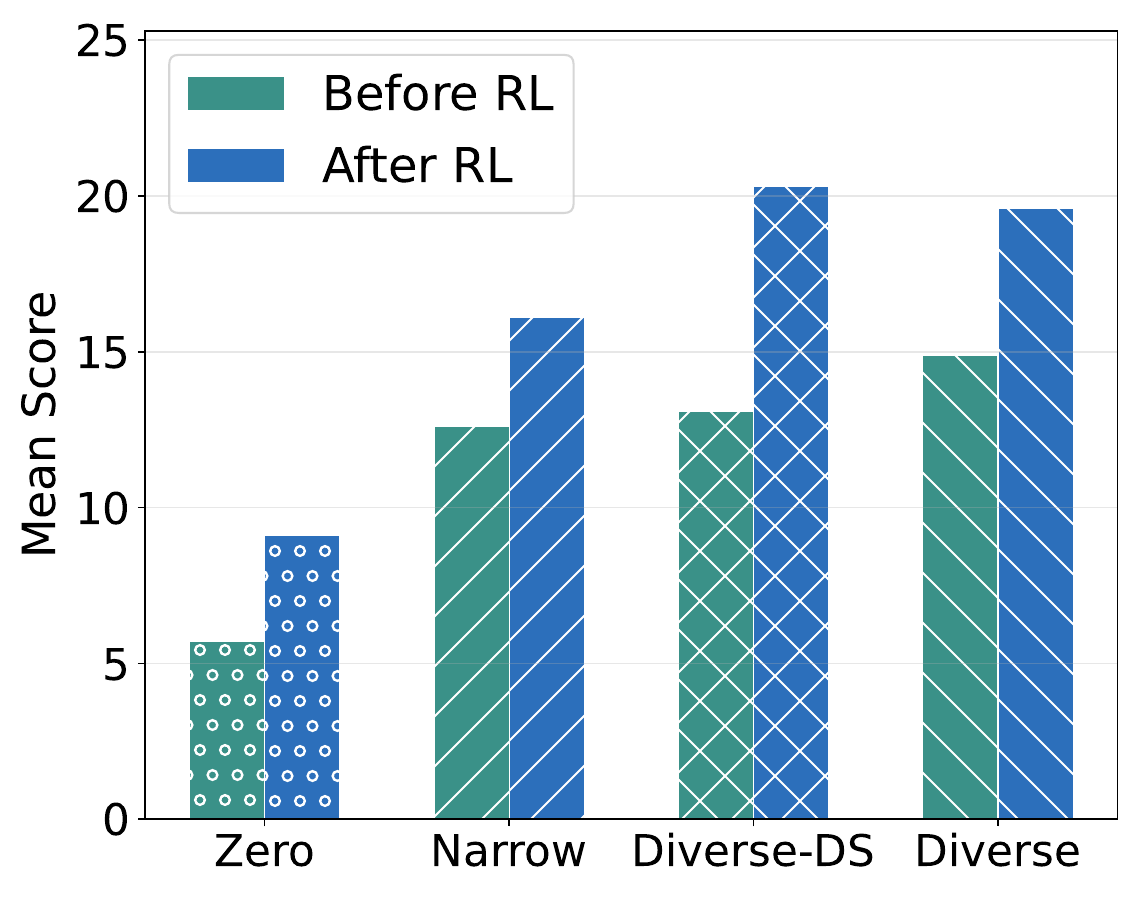}
    \caption{
    % SFT ablations. 
    Mean score over the four evaluation benchmarks Before and after RL on TMax-RIVER-3.5K, for Base model, Narrow-SFT, Diverse-DS-SFT, and Diverse-SFT.
    % \textcolor{red}{TODO: adding experiments in Qwen3-4B.}
    }
    % \vspace{-10pt}
    \label{fig:ablation-different-sft}
\end{wrapfigure}

% As introduced in Alg.~\ref{alg:recipe}, we perform warm-start SFT followed by RL on the TMax-\method-3.5K environments. 
The average performance across the four benchmarks (TWV, TB-v2.1, TB-Pro, and OTBL) of the checkpoints before and after RL for these four initialization settings is presented in Fig.~\ref{fig:ablation-different-sft}. 
We observe that, without SFT, performing RL directly on the base model leads to the worst evaluation performance, motivating the use of SFT before RL for relatively weak models such as Qwen3-8B.
Comparing \textbf{Narrow-SFT} with \textbf{Diverse-DS-SFT}, although the two achieve comparable performance after SFT, a substantial gap remains after RL, indicating the importance of acquiring diverse atomic skills before RL for improved generalizability.

\begin{figure}[h]
  \centering
  \includegraphics[width=\linewidth]{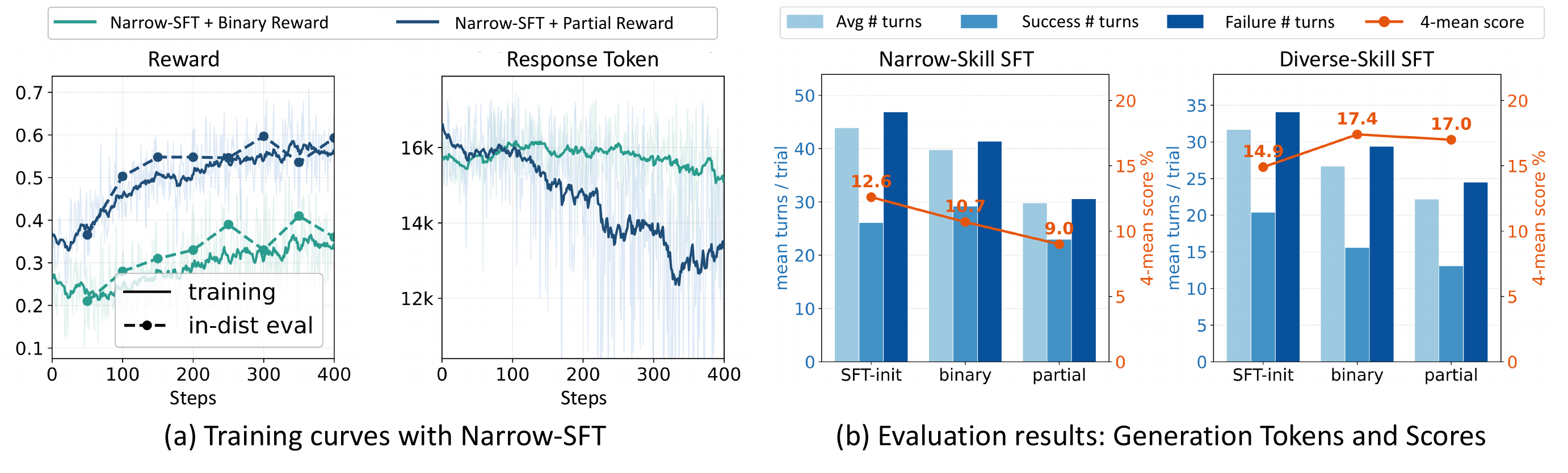}
  \vspace{-10pt}
  \caption{Training curves and evaluation results for different SFT initializations ({Narrow-SFT} vs. {Diverse-SFT}) and trajectory-level reward designs ({binary} vs. {partial}). All RL experiments are conducted using TermiGen~\citep{zhu2026termigen}.}
  % \vspace{-10pt}
  \label{fig:reward-hacking}
\end{figure}

In another set of experiments, we further observe that \textbf{Diverse-SFT} can reduce reward hacking. During an early stage of our exploration of RL with environments from TermiGen~\citep{zhu2026termigen}, we adopted partial rewards defined as $r_p=0.5\times$ pass-rate $+\ 0.5\times$ full-pass-bonus. The training curves are presented in Fig.~\ref{fig:reward-hacking}~(a), where both the training rewards and in-distribution evaluation scores improve during RL. However, this setup introduces two potential sources of reward hacking: (1) \textbf{Hackable verifiers}: the environments had not yet been filtered by our \method\ pipeline. The evaluation results in Fig.~\ref{fig:env-audit-donuts} (App.~\ref{app:env-audit}) indicate that this dataset is not sufficiently clean for RL\footnote{TermiGen environments were used for SFT data collection in their experiments.}; and (2) \textbf{Early termination}: under partial rewards, the agent may learn a shortcut that satisfies only a subset of the verifiers and then prematurely claims task completion. As shown in Fig.~\ref{fig:reward-hacking}~(b, left), RL degrades generalizability even as in-distribution performance improves, which is consistent with reward hacking. Combined with the decrease in generation length during training and our trajectory analysis, these results indicate that this form of reward hacking primarily stems from early termination.

Using \textbf{Diverse-SFT} mitigates these reward-hacking risks. As shown in Fig.~\ref{fig:reward-hacking}~(b, right), when starting from Diverse-SFT while keeping all other configurations unchanged, the evaluation performance after RL improves under both the binary-reward and partial-reward settings. Our interpretation is that exposure to diverse atomic skills equips the model with the capabilities needed to solve the training tasks, making it less likely to exploit shortcuts during policy optimization and thereby mitigating reward hacking to some extent.

\begin{AIbox}{Section~\ref{subsection:sft-analysis} Takeaways}
$\bullet$ For relatively weak base models (e.g., Qwen3-8B), SFT with diverse atomic-skill coverage provides an important warm start for RL to improve generalizability;\\
$\bullet$ SFT with broader skill coverage also helps reduce reward hacking.
\end{AIbox}

\subsection{RL Analysis}
\label{subsection:rl-analysis}
\begin{wrapfigure}{r}{0.65\linewidth}
    \centering
    \vspace{-25pt}
    \includegraphics[width=\linewidth]{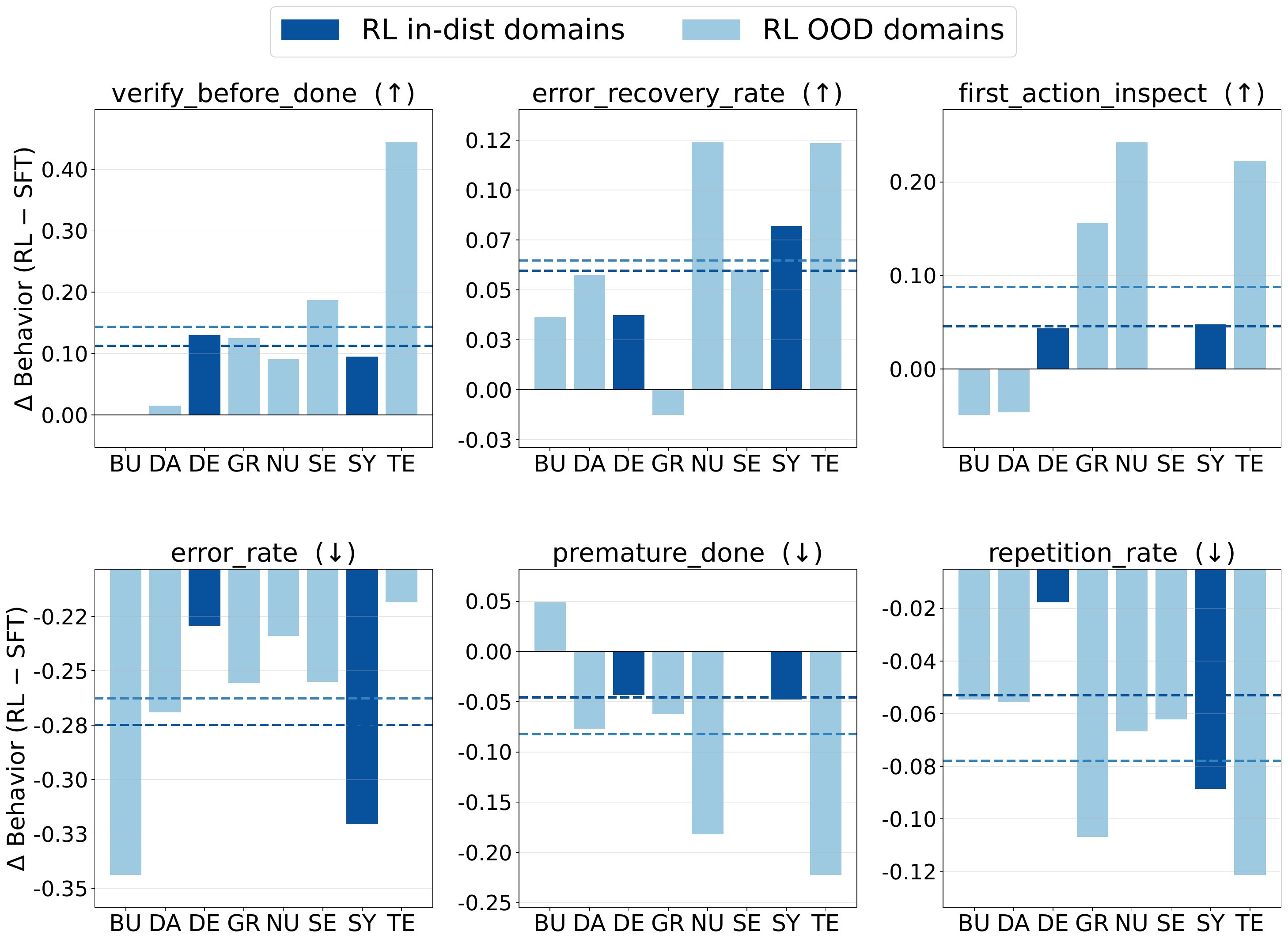}
    \caption{
    Behavior differences after RL (post-RL ratio - pre-RL ratio). Domains: BU = Build, DA = Data, DE = Debug, GR = Graph, NU = Numerical, SE = Security, SY = Systems, TE = Text. $\uparrow$/$\downarrow$: the higher/lower, the better from empirical evaluations.}
    \vspace{-10pt}
    \label{fig:RL-subset-exp}
\end{wrapfigure}

In this section, we conduct more controlled experiments to better understand generalization in terminal-agent RL. We divide the \method-TMax-3.5K dataset into $8$ categories based on the required domain skills, as shown in Tab.~\ref{tab:indist_domain}, and construct a held-out evaluation set of $\sim300$ tasks from \method-TMax-3.5K. We then perform RL on different subsets of domains, ranging from all $8$ domains down to only $2$. For RL (2 doms), training uses only $\sim800$ environments from the Debug and System domains. The evaluation results of the RL-trained models and the SFT model on the held-out set are presented in Tab.~\ref{tab:indist_domain}. We observe that reducing the number of training domains to $4$ does not decrease overall performance, and we find no significant performance degradation even on specific domains excluded from training. When the number of training domains is reduced to $2$, such that the training environments cover only Debug and System, overall performance decreases but remains substantially higher than that of the SFT model. Moreover, performance on untrained and OOD domains, such as Data and Numerical, still improves.

We also visualize $6$ representative behaviors in the evaluation traces and measure the changes in their prevalence before and after RL for the $2$-domain setting. The results are presented in Fig.~\ref{fig:RL-subset-exp}. We observe that these behaviors are shaped not only in the trained domains (darker bars), but also in the untrained domains (lighter bars). This suggests that the behavioral changes induced by RL generalize beyond the training domains and contribute to performance improvements in other domains.

\newcommand{\gain}[2]{#1~{\scriptsize(+#2)}}
\begin{table}[ht]
\centering
\small
\setlength{\tabcolsep}{4pt}
\resizebox{1.\linewidth}{!}{
\begin{tabular}{lccccccccc}
\toprule
Category & Build & Data & Debug & Graph & Numerical & Security & Systems & Text & ALL \\
\midrule
RL (8 doms) & \gain{42.6}{8.2} & \gain{56.9}{27.7} & \gain{54.3}{10.9} & \gain{68.8}{40.6} & \gain{69.7}{24.2} & \gain{40.6}{18.8} & \gain{52.4}{26.2} & \gain{55.6}{11.1} & \gain{53.9}{21.5} \\
RL (6 doms) & \gain{45.9}{11.5} & \gain{50.8}{21.5} & \gain{58.7}{15.2} & \gain{65.6}{37.5} & \gain{63.6}{18.2} & \gain{43.8}{21.9} & \gain{61.9}{35.7} & \gain{44.4}{0.0} & \gain{55.2}{22.9} \\
RL (4 doms) & \gain{50.8}{16.4} & \gain{53.8}{24.6} & \gain{47.8}{4.3} & \gain{68.8}{40.7} & \gain{69.7}{24.2} & \gain{40.6}{18.7} & \gain{51.2}{25.0} & \gain{55.6}{11.2} & \gain{53.6}{21.3} \\
RL (2 doms) & \gain{45.9}{11.5} & \gain{47.7}{18.5} & \gain{52.2}{8.7} & \gain{53.1}{25.0} & \gain{57.6}{12.1} & \gain{46.9}{25.0} & \gain{46.4}{20.2} & \gain{55.6}{11.1} & \gain{49.2}{16.9} \\
SFT & 34.4 & 29.2 & 43.5 & 28.1 & 45.5 & 21.9 & 26.2 & 44.4 & 32.3 \\
\bottomrule
\end{tabular}
}
\caption{Solve rate (\%) on the held-out evaluation set by task
domain.}
\label{tab:indist_domain}
\end{table}

\begin{AIbox}{Section~\ref{subsection:rl-analysis} Takeaways}
$\bullet$ RL shapes behaviors beyond the training domains and fosters generalizability.
\end{AIbox}

\section{Conclusion}
% \section{Conclusion}
\label{section:conclusion}
In this work, we study how RL improves terminal agents and propose the \textit{agentic compositional generalization} hypothesis. Our empirical findings support complementary roles for different stages of model training: pre-training and SFT provide low-level atomic skills, while RL primarily shapes high-level decision-making behaviors that compose, route, and coordinate these skills across domains. This perspective is consistent with the strong OOD generalization of agentic RL and highlights \emph{verifier quality}, which determines the training signal that shapes these behaviors, as a key factor beyond simply scaling the quantity or diversity of training environments. Guided by this insight, we introduce \method, a simple training recipe that improves the integrity of the reward signal by filtering low-quality environments and optionally incorporating behavior-level regularization. \method\ achieves the best performance among evaluated RL-trained 8B models across four terminal-agent benchmarks. Moreover, \method\ generalizes across model families, model scales, agent harnesses, and RL objectives: using fewer than $30\%$ of the TMax training environments, we achieve, on average, $106\%$ and $30\%$ larger RL gains on Terminal-Bench-Lite and Terminal-Bench-v2.1, respectively, across models ranging from $2$B to $27$B. Together, our results suggest that improving the quality of training signals, rather than scaling environment coverage alone, provides an effective and efficient path toward building more generalizable terminal agents.

% \textbf{Future Works.}\\
% (1) Discovering behaviors with better casual relationship with performance; \\
% (2) Investigating how to improve the general quality of synthetic environments;

\section*{Acknowledgment}
We gratefully acknowledge the authors of the TMax collection for their valuable contribution to the community. Their release provides an important resource for advancing research on RL for terminal agents and was instrumental to this work. We also thank Jiacheng Zhu and Zhepeng Cen for their helpful discussions.

\bibliography{iclr2027_conference}
\bibliographystyle{iclr2027_conference}

\clearpage
\tableofcontents
\clearpage

\appendix

% \textcolor{red}{TODO: content}
\section{Training Details}
\label{app:training-details}

% ---------------------------------------------------------------------
% Supervised fine-tuning
% ---------------------------------------------------------------------
\subsection{Supervised Fine-Tuning}
\label{app:sft-training}

\paragraph{Training data.}
We finetune on multi-turn terminal-agent trajectories in an
\emph{endless-agent} format~\cite{gandhi2026endless}: a system prompt, the task instruction, and an
alternating sequence of assistant turns and environment observations, where
every assistant turn contains an explicit chain-of-thought
(\texttt{<think>...</think>}) followed by a single shell command or a
terminating action. We obtained \textbf{Diverse-SFT} with $78$K trajectories after format converting and filtering. For more SFT dataset discussions and atomic skill coverage analysis, please refer to App.~\ref{appendix-sub:sft-data-coverage}.
% We use two datasets that share this format but differ in scale:
% \begin{itemize}
%   \item \textbf{OT-78K} (\emph{full set}): $\sim$78K trajectories, our largest
%         SFT corpus.
%   \item \textbf{DS-13600}: a uniform-random subsample of OT-78K to
%         $13{,}600$ trajectories (fixed seed), used as a size-matched control.
%         It is drawn i.i.d. from OT-78K, so its skill/domain distribution is
%         statistically identical to the full set; the only variable it changes
%         relative to OT-78K is \emph{data volume}, which lets us isolate the
%         effect of dataset size from that of data composition.
% \end{itemize}
% All trajectories are truncated to a maximum context length of $16{,}384$
% tokens, which covers $\geq 99\%$ of examples (the $99$th-percentile
% trajectory is $\approx16.5$K tokens).

\paragraph{Model and optimization.}
We perform {full-parameter} fine-tuning of \textbf{Qwen3-8B} using the LLaMA-Factory framework~\citep{zheng2024llamafactory}. Loss is computed on assistant tokens (generated tokens) only. Key
hyperparameters are listed in Table~\ref{tab:sft-hparams}; experiments with both datasets are
done with the identical recipe, so any downstream difference is
attributable to the data alone.

\begin{table}[h]
\centering
\small
\caption{SFT hyperparameters.}
\label{tab:sft-hparams}
\begin{tabular}{@{}ll@{}}
\toprule
\textbf{Hyperparameter} & \textbf{Value} \\
\midrule
Base model              & Qwen3-8B \\
Fine-tuning type        & Full parameter \\
Precision               & bf16 \\
Epochs                  & 1 \\
Learning rate           & $1\times10^{-5}$ \\
LR schedule             & Cosine \\
Warmup ratio            & 0.1 \\
Effective batch size    & 64 \\
\quad(per-device $\times$ GPUs $\times$ grad.\ accum.) & ($1\times8\times8$) \\
Max sequence length     & 16{,}384 \\
Optimizer sharding      & DeepSpeed ZeRO-2 \\
Attention               & SDPA \\
\bottomrule
\end{tabular}
\end{table}

% ---------------------------------------------------------------------
% Reinforcement learning
% ---------------------------------------------------------------------
\subsection{Reinforcement Learning}
\label{app:rl-training}
This section provides details for training our $8$B models.
\paragraph{Training environments.}
For full \method, we train on the \textbf{\method-TMax-3.5K} environment set. 

\paragraph{Training method.}
For the {Qwen3} model family, We fine-tune the policy with \textbf{Group Relative Policy Optimization (GRPO)}~\citep{shao2024deepseekmath}. For each prompt we sample a group of $G$ rollouts from the current policy, execute them in the terminal environment, and score each rollout with the task verifier under a \emph{sparse binary} reward (\,$r\in\{0,1\}$\,: $1$ iff the verifier passes). The advantage term is defined and discussed in (\ref{eq:shaped-adv}).

\paragraph{Objective.}
Let $\rho_t(\theta)=\pi_\theta(a_t\mid s_t)/\pi_{\theta_{\text{old}}}(a_t\mid s_t)$
be the per-token importance ratio and $A_t$ the (shaped) token advantage.
% We optimize a clipped policy objective with an explicit KL regularizer to a
% frozen reference policy $\pi_{\text{ref}}$:
\begin{equation}
  \mathcal{L}(\theta) \;=\;
  -\,\mathbb{E}_t\!\Big[
      \min\!\big(\rho_t A_t,\;
      \operatorname{clip}(\rho_t,\,1-\varepsilon_{\text{low}},\,1+\varepsilon_{\text{high}})\,A_t\big)
    \Big]
  \;+\; \beta\,\mathrm{KL}\!\big(\pi_\theta \,\|\, \pi_{\text{ref}}\big),
  \label{eq:grpo-loss}
\end{equation}
with asymmetric clipping bounds $\varepsilon_{\text{low}}$, and $\varepsilon_{\text{high}}$. The per-token losses are aggregated with a
sequence-mean reduction.

% \paragraph{Advantage shaping: repetition penalty.}
% On top of the outcome advantage in Eq.~\ref{eq:grpo-adv}, we apply a local
% \textbf{repetition penalty} as a per-token advantage correction. Within a
% rollout we detect \emph{harmful repetition loops} --- a command that is
% re-issued and returns the same (non-progressing) observation many times ---
% and subtract a fixed penalty $\delta$ from the advantage of the offending
% response tokens, leaving the trajectory-level advantage otherwise intact:
% % \begin{equation}
% %   \tilde{A}_t \;=\; A_t \;-\; \delta \cdot \mathbb{1}\!\left[\,t \in \mathcal{P}\,\right],
% %   \label{eq:rep-penalty}
% % \end{equation}
% \begin{equation}
%   \widetilde{A}_t
%   :=
%   A_t
%   -
%   \delta\,\mathbf{1}\!\left\{t \in \mathcal{P}\right\},
%   \label{eq:rep-penalty}
% \end{equation}
% where $\mathcal{P}$ is the set of response tokens belonging to penalized
% (looping) turns. An optional per-rollout cap bounds the total subtraction so
% that a heavily looping trajectory cannot dominate the outcome signal. The
% penalty is applied \emph{after} advantage normalization.

\paragraph{Hyperparameters.}
We adopt most hyperparameters for RL training from the endless-terminal work~\citep{gandhi2026endless}.
Table~\ref{tab:rl-hparams} lists the training hyperparameters. 

\paragraph{Computation details.}
All RL runs for Qwen3-8B models use a single such node.
Each node is an 8×NVIDIA H200 GPU server, with $\sim$143 GB HBM3e per GPU ($\sim$1.15 TB aggregate GPU memory), fully NVLink-connected. Host side: 2× Intel Xeon Platinum 8488C (96 logical cores), and $\sim$2 TB system RAM. For the main experiments, we run for around $60$ hours to perform $400$ RL steps.

\begin{table}[h]
  \centering
  \caption{RL (GRPO) training hyperparameters.}
  \label{tab:rl-hparams}
  \begin{tabular}{@{}ll@{}}
    \toprule
    \textbf{Hyperparameter} & \textbf{Value} \\
    \midrule
    \multicolumn{2}{@{}l}{\emph{Optimization}}\\
    Base policy                       & Qwen3-8B (post-trained checkpoint) \\
    Advantage estimator               & GRPO \\
    Learning rate                     & $5\times10^{-7}$ \\
    Warmup steps                      & $20$ \\
    Weight decay                      & $0.01$ \\
    % Max gradient norm                 & $0.1$ \\
    Training epochs                   & $10$ \\
    Update epochs per batch           & $2$ \\
    Train batch size (prompts)        & $32$ \\
    Policy mini-batch size            & $32$ \\
    Parameter sharding                & FSDP2, 1 node $\times$ 8 GPUs \\
    Max RL Step                       & 400 \\
    \midrule
    \multicolumn{2}{@{}l}{\emph{Objective}}\\
    % KL loss coefficient $\beta$       & $0.001$ \\
    KL in reward                      & disabled \\
    Clip range (low, high)            & $(0.2,\ 0.28)$ \\
    Loss reduction                    & sequence-mean \\
    Reward                            & sparse binary $\{0,1\}$ \\
    \midrule
    \multicolumn{2}{@{}l}{\emph{Rollout / generation}}\\
    Samples per prompt            & $8$ \\
    Max turns per episode             & $16$ \\
    Max prompt length                 & $1536$ \\
    Max input length                  & $16384$ \\
    Max generation length (per turn)  & $2048$ \\
    Sampling temperature              & $0.6$ \\
    Top-$p$                           & $1.0$ \\
    Inference engines                 & $8$ \\
    \midrule
    \multicolumn{2}{@{}l}{\emph{Repetition penalty}}\\
    Penalty $\delta$ (per token)      & $0.05$ \\
    Per-rollout cap                   & $0.15$ \\
    \bottomrule
  \end{tabular}
\end{table}

% ---------------------------------------------------------------------
% Evaluation
% ---------------------------------------------------------------------
\subsection{Evaluation}
\label{app:evaluation}

\paragraph{Serving.}
All checkpoints are evaluated under an identical serving configuration. We convert each policy checkpoint to Hugging Face \texttt{safetensors} and serve it with \textbf{SGLang}~\citep{zheng2024sglang}. A single 8-GPU node
hosts one model instance with tensor parallelism $\text{TP}{=}8$ (we do {not} use data-parallel replicas, which we found to deadlock under the agent's synchronous per-turn calls) and a static KV-cache memory fraction of $0.85$. The served context length is the model's full $40{,}960$-token window (\texttt{max\_position\_embeddings}). 

\paragraph{Decoding and budgets.}
Rollouts use temperature $T{=}0.6$ and top-$p{=}1.0$, with up to $2048$ tokens generated per turn. Each episode is capped at $64$ action turns and a $600$-second wall-clock budget, whichever comes first; a task that reaches either cap without a terminating action is scored as a failure.

\paragraph{Evaluation Benchmarks.}
We evaluate on four terminal-agent benchmarks and report their mean score, summarized in Tab.~\ref{tab:eval-benchmarks}. All benchmarks are in the {harbor}~\citep{harbor2026} task format. For \textbf{OpenThoughts-TBLite}, the score is a continuous value per their design. For the rest benchmarks, scoring is {binary pass/fail} per task (reward $1$ iff the verifier's tests pass).

\begin{table}[h]
\centering
\small
\caption{Evaluation benchmarks. The averaged score is the unweighted mean of the four test benchmarks.}
\label{tab:eval-benchmarks}
\begin{tabular}{@{}llll@{}}
\toprule
\textbf{Benchmark} & \textbf{\#Tasks} & \textbf{Score Type} & \textbf{Focus} \\
\midrule
OpenThoughts-TBLite      & $100$ & Continuous & General CLI / file-ops / SWE \\
Terminal-Bench~2.1       & $89$  & Binary & SWE, ML, security, data science \\
Terminal-Bench-Pro       & $200$ & Binary & SWE, systems, debugging, data \\
TerminalWorld-Verified   & $200$ & Binary & Mixed (categorized, verified) \\
\bottomrule
\end{tabular}
\end{table}

\begin{table}[h]
\centering
\small
\caption{Evaluation serving and decoding configuration, held fixed across all
checkpoints and benchmarks.}
\label{tab:eval-config}
\begin{tabular}{@{}ll@{}}
\toprule
\textbf{Setting} & \textbf{Value} \\
\midrule
Serving engine            & SGLang \\
Tensor parallelism        & $8$ (\texttt{--tp 8}) \\
KV-cache mem.\ fraction   & $0.85$ \\
Served context length     & $40{,}960$ tokens \\
Sampling temperature      & $0.6$ \\
Top-$p$                   & $1.0$ \\
Max tokens per turn       & $2048$ \\
Max turns per episode     & $64$ \\
Wall-clock budget / task  & $600$\,s \\
Task concurrency          & $8$ \\
\bottomrule
\end{tabular}
\end{table}

\clearpage
\newpage
\section{Data and Environment Details}
\subsection{SFT dataset coverage}
% \textcolor{red}{TODO: separate section}
\label{appendix-sub:sft-data-coverage}
\begin{figure}[h]
  \centering
  \includegraphics[width=0.86\textwidth]{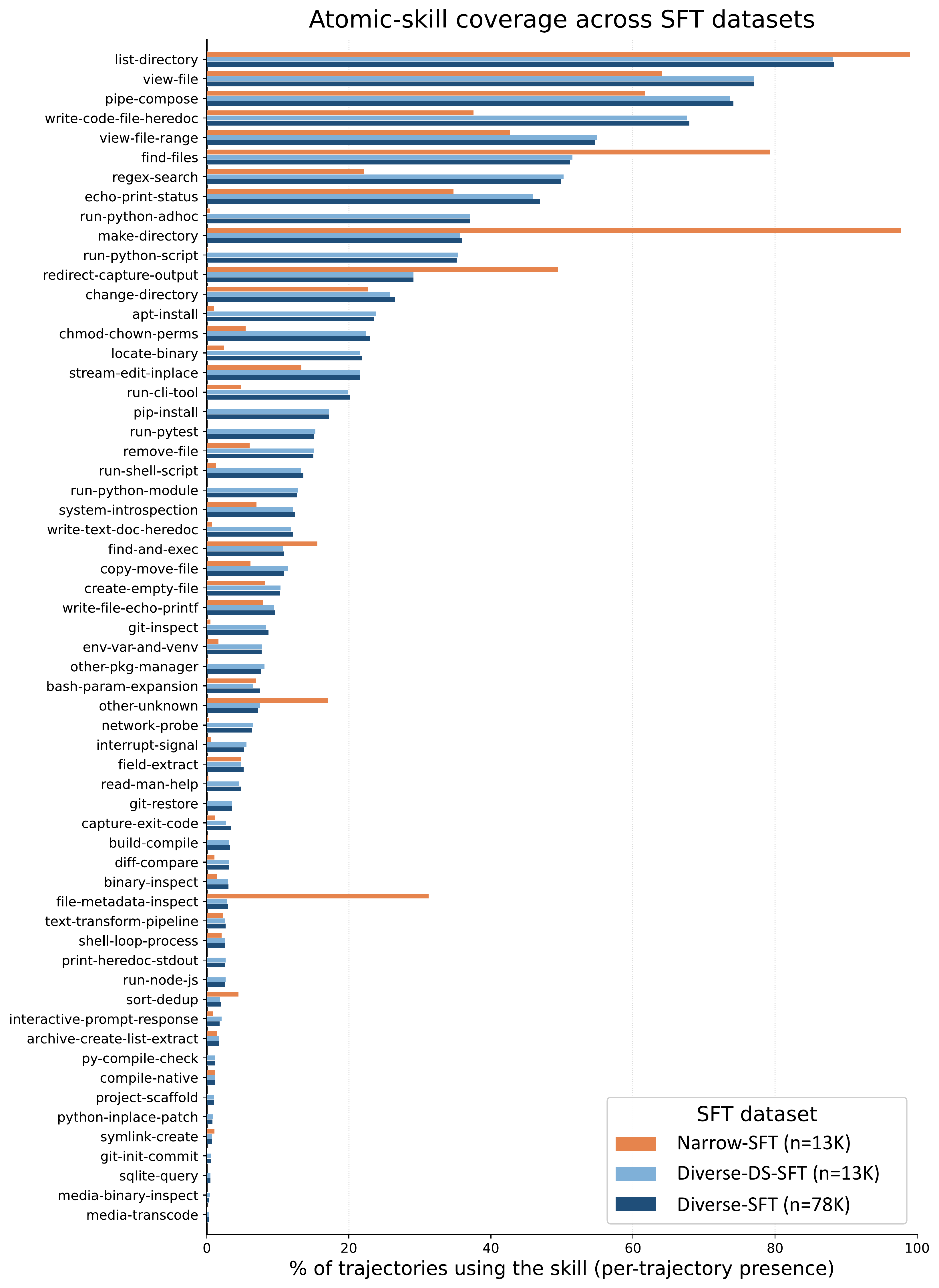}
  \caption{Atomic-skill coverage of the three SFT datasets, measured as the
fraction of trajectories labeled by GPT-5.4 as exhibiting each of the
$60$ command-level skills.}
  \label{fig:atomic-coverage-3way}
\end{figure}

\paragraph{SFT Construction Details.}
The source SFT data OpenThoughts-Collection~\citep{guha2026openthoughts, raoof2026openthoughts} uses the \textbf{Terminus-2} harness, whereas our stack consumes the \textbf{EndlessAgent}~\citep{gandhi2026endless} harness, so we translate every trajectory deterministically and drop any that cannot be represented. 
In \textbf{Terminus-2}, the content in one generation turn is a JSON object with \texttt{analysis}, \texttt{plan}, a \textbf{batch} of \texttt{commands}, and \texttt{task\_complete}, and each observation is a full terminal feedback. In EndlessAgent, the content in one generation turn consists of the following content: $\texttt{<think>}\dots\texttt{</think>}\,\texttt{<command>}c\texttt{</command>}$ for one command $c$ (or $\texttt{<action>done</action>}$ to claim the task is done), and observations use the fixed template \texttt{Command executed successfully. Output: <stdout> (exit\_code=0)}.

The translation (1) replaces the system prompt with the EndlessAgent one; (2) extracts the task from the first user message (the span after
\texttt{Task Description:}); (3) merges a turn's command batch $[c_1,\dots,c_k]$ into $\texttt{<command>}c_1\,\texttt{\&\&}\cdots\texttt{\&\&}\,c_k\texttt{</command>}$; (4) fills \texttt{<think>} from \texttt{analysis}+\texttt{plan}; and (5) reduces each raw terminal screen to the command's actual output by removing the shell prompt and the echo of the typed command (including the
continuation lines of multi-line commands), while carefully preserving genuine output that happens to start with ``\texttt{>}'' (e.g.\ \texttt{diff} results).

We drop a trajectory if its first message lacks a \texttt{Task Description:} marker, or if any assistant turn \emph{other than the last} fails to parse as valid JSON. A non-JSON turn appearing \emph{as the final turn} is not treated as corruption: it is the terminus ``are you sure?'' confirmation, which we convert to \texttt{done}. Because the trainer silently discards any trajectory with two consecutive assistant turns, we treat \texttt{done} as terminal and splice a synthetic observation wherever one assistant turn would directly follow another;
The remaining trajectories all end in an explicit \texttt{done}.

% \textcolor{red}{TODO: SFT construction details}
\paragraph{Skill composition of the SFT datasets.} Fig.~\ref{fig:atomic-coverage-3way} compares the atomic skill
composition of the three SFT corpora ({Narrow-SFT}, {Diverse-DS-SFT}, and {Diverse-SFT}) used in our study, labeling every trajectory with a $60$-skill atomic taxonomy (via GPT-5.4) and reporting per-skill coverage. 
The comparison serves two purposes. First, it validates {Diverse-DS-SFT} as a clean {size} control for {Diverse-SFT}: because {Diverse-DS-SFT} is drawn i.i.d.\ from {Diverse-SFT},
its coverage is indistinguishable from the full set across all $60$ skills, so any downstream difference between models trained on {Diverse-SFT} and {Diverse-DS-SFT} is attributable to data \emph{volume} rather than to a shift in skill composition. Second, it isolates {what} distinguishes {Narrow-SFT} from {Diverse-SFT}: the gap is not a uniform thinning but a categorical one on the
execution-oriented skills. Narrow-SFT is dominated by filesystem navigation and authoring yet contains almost no code execution or environment setup
(e.g., \texttt{run-python}, \texttt{run-pytest}, \texttt{pip}/\texttt{apt-install} all near $0\%$), whereas {Diverse-SFT} and {Diverse-DS-SFT} exercise these skills with SFT.

\subsection{RL environments}

\begin{figure}[ht]
  \centering
  \includegraphics[width=\textwidth]{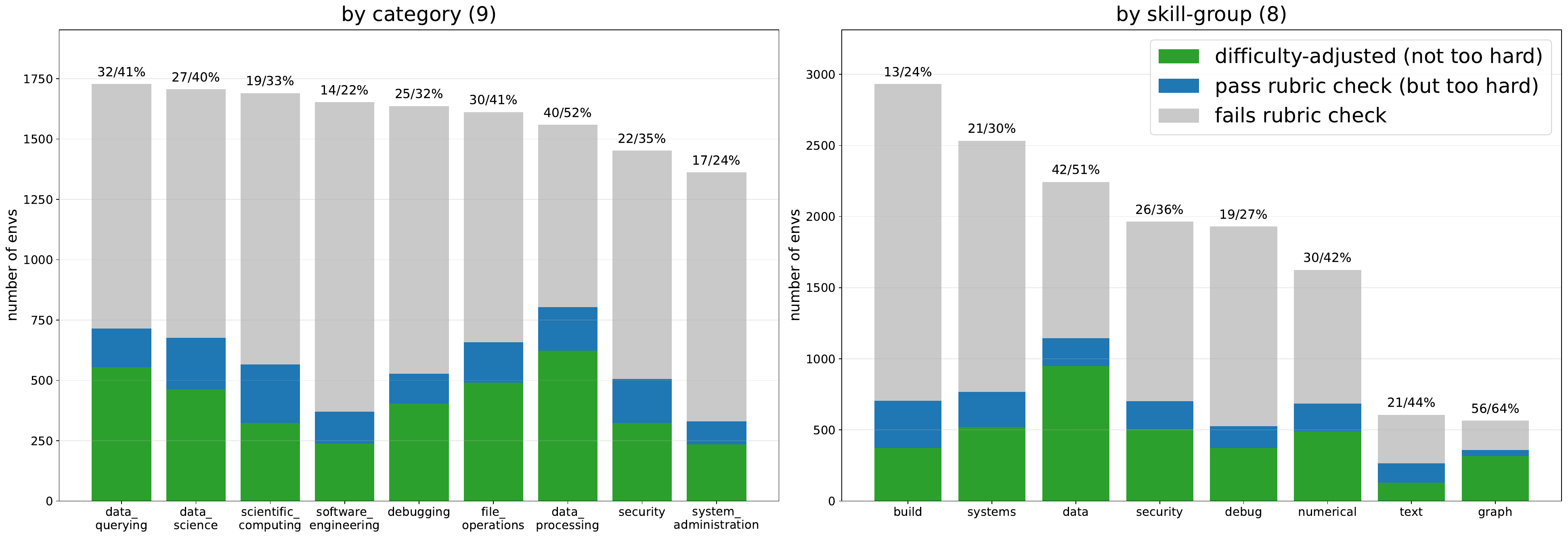}
  \caption{%
    {Environment quality-and-difficulty funnel for the TMax training pool
    (14{,}399 environments), broken down by task \emph{category} (left, 9 classes)
    and by \emph{skill group} (right, 8 groups).}
  }
  \label{fig:env-quality-funnel}
\end{figure}

In Fig.~\ref{fig:env-quality-funnel}, each environment is passed through a two-stage filter, and every bar shows the
    outcome as a stacked count:
    {(1)} \textbf{fails rubric check} (grey): flagged by our rubrics and GPT-5.4 judges;
    {(2)} \textbf{pass rubric check but too hard} (blue): the environment
    is clean, but a capable reference agent cannot solve it;
    {(3)} \textbf{difficulty-adjusted} (green): clean \emph{and} solvable
    within the budget. These environments consist of our \textbf{\method-TMax-3.5K}.
    The annotation above each bar reports \texttt{difficulty-adjusted\% / rubric-pass\%} of the original pool.
    Categories are the task-provided domain labels; skill groups are assigned by
    mapping each environment's primary skill onto one of eight groups ({build, data, debug, graph, numerical, security, systems,
    text}).

\clearpage
\newpage
\clearpage
\newpage
\section{Environment Evaluation}
\label{app:env-audit}
\begin{figure*}[t]
  \centering
  \includegraphics[width=\textwidth]{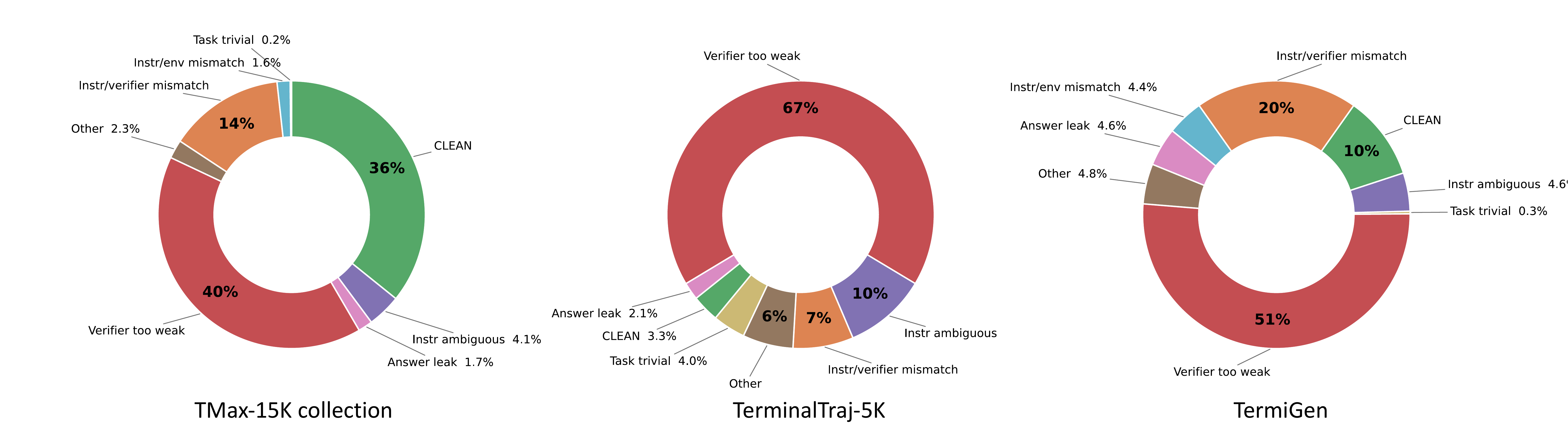}
  \caption{%
    Rubrics evaluation results conducted by GPT-5.4: (left to right): Environments from TMax~\citep{ivison2026tmax}; environments from TerminalTraj~\citep{wu2026large}; and  environments from TermiGen~\citep{zhu2026termigen}).
  }
  \label{fig:env-audit-donuts}
\end{figure*}

We perform evaluations on three candidate environment pools using our environment-filtering rubrics. We emphasize that our evaluations may contain both false positives and false negatives. The results (Fig.~\ref{fig:env-audit-donuts}) differ sharply across pools. \textbf{TMax} is the cleanest, with $35.8\%$ \textsc{Clean}, but still contains $40.4\%$ of environments flagged as \textsc{Verifier-too-weak}. \textbf{TermiGen} is only $10.1\%$ \textsc{Clean} ($51.4\%$ too-weak, $19.6\%$ mismatch), while \textbf{TerminalTraj-5k} performs the worst under our rubrics, with only $3.3\%$ \textsc{Clean} and $67.2\%$ \textsc{Verifier-too-weak}. Across all three pools, the dominant failure mode is weak verification (\textsc{Verifier-Too-Weak}). We next provide the full rubric (Sec.\ref{app:env-audit-rubric}) and show representative examples of verdicts (Sec.\ref{app:env-audit-examples}).

\subsection{Evaluation Rubrics and Issue Categories}
\label{app:env-audit-rubric}
The evaluation agent reads the three artifacts that must be mutually consistent: \texttt{instruction.md} (the task specification presented to the agent), the reward \emph{verifier} (the pytest that scores a rollout), and the \emph{environment} (the Dockerfile and any bundled data). It then asks a single agent-centric question: \emph{would a competent agent that faithfully follows the instruction be able to (a) understand the task unambiguously and (b) actually pass the verifier?} Each task is assigned exactly one of eight verdicts:

\begin{itemize}
  \item \textbf{\textsc{Clean}:} Instruction is clear and self-sufficient, a correct-per-instruction solution passes, AND the verifier genuinely validates correctness: it recomputes/decodes the true answer from the input (or exact-compares against a correct hardcoded answer) rather than merely checking output shape. A hard-but-fair task is still \textsc{Clean}.
  \item \textbf{\textsc{Verifier-Too-Weak}:} The verifier passes \emph{without} the task actually being solved: it checks only output shape/format/types/count, trusts a self-reported value, greps the source for keywords, checks a hardcoded subset, or tests only fixed inputs so the answer can be hardcoded. A fabricated, plausibly-shaped output would pass. 
  \item \textbf{\textsc{Instr-Verifier-Mismatch}:} The verifier checks something the instruction does not imply, or contradicts it. For example, reads/writes a different path, asserts a hardcoded value that conflicts with the data, or demands a stricter format. A solution that faithfully follows the instruction still fails.
  \item \textbf{\textsc{Instr-Env-Mismatch}:} The instruction or verifier references input paths, files, dependencies, or services the environment does not actually provide, so the task cannot be completed as written (missing data/compiler/service).
  \item \textbf{\textsc{Answer-Leak}:} The instruction itself, or a file shipped in the environment, reveals the expected answer (lists the result, embeds the solution, or provides a ground-truth file), so the task requires no genuine work.
  \item \textbf{\textsc{Instr-Ambiguous}:} The instruction alone is underspecified: a \emph{necessary} specification is missing, so two equally reasonable interpretations produce different outputs and the verifier accepts only one. (Merely admitting multiple valid solution \emph{paths} is not enough.)
  \item \textbf{\textsc{Task-Trivial}:} The task itself is so easy there is no meaningful work or learning signal: write a constant, echo a value, copy a file, regardless of verifier strength.
  \item \textbf{\textsc{Other}:} A genuine defect fitting none of the above: verifier code that is buggy/crashes; a task needing network/GPU/GUI/hardware the container lacks; non-determinism; a required background service that is never started; or an internally contradictory spec.
\end{itemize}

\subsection{Cross-model validation of the audit labels}
\label{app:env-audit-validation}

The verdicts above were produced by GPT-5.4. To assess whether these judgments are robust rather than idiosyncratic to a single model, we conducted an independent blind re-judgment using a different model, Claude Opus 4.8~\citep{anthropic2026opus48}. We constructed a stratified sample of $120$ TMax environments, with $15$ examples drawn from each of the eight GPT-5.4 verdict categories. To avoid anchoring, each environment was stripped of its GPT-5.4 label and any category cue, shuffled, and provided to a dedicated judge that independently read the full instruction, Dockerfile, and verifier code before assigning a verdict. The two labels were compared only afterward.

At the fine-grained level, the two models assign the same eight-way verdict to only $22.5\%$ of the $120$ examples. This disagreement does not necessarily imply that one judgment is incorrect: a single defective environment can exhibit multiple failure modes simultaneously. For example, a verifier that hard-codes its expected output may reasonably be categorized as \textsc{Verifier-too-weak}, while the same environment may also exhibit characteristics of \textsc{Instr/verifier-mismatch} or \textsc{Task-trivial}. Thus, forcing each environment into a single primary category can lead the two models to disagree on \emph{which} defect is most salient even when both identify the environment as defective.

For training, the more consequential distinction is binary: whether an environment is usable (\textsc{Clean}) or exhibits at least one defect. Under this binary grouping, the two models agree on $100/120$ examples ($83.3\%$) in our stratified sample, as summarized in Tab.~\ref{tab:audit-crossmodel}. Because the sample was stratified by the GPT-5.4 verdict rather than drawn according to the natural label distribution of the full pool, this number should be interpreted as agreement under our sampling design rather than as an estimate of population-level agreement.

More informative are the conditional agreement rates. Among the $105$ examples labeled defective by GPT-5.4, the independent judge also identifies $94$ as defective ($89.5\%$). In contrast, among the $15$ examples labeled \textsc{Clean} by GPT-5.4, only $6$ are independently judged \textsc{Clean} ($40.0\%$). This asymmetry suggests that defect labels are substantially more stable across judges than \textsc{Clean} labels. It also indicates that our reported clean rates may be optimistic, since some environments labeled \textsc{Clean} by GPT-5.4 are judged defective by the independent model. Overall, this validation supports using the audit as a conservative filtering mechanism for removing environments with readily detectable defects, while treating fine-grained defect categories and \textsc{Clean} labels with greater caution.

\begin{table}[t]
\centering
\small
\begin{tabular}{lcc}
\toprule
GPT-5.4 label & \multicolumn{2}{c}{Independent blind re-judge} \\
\cmidrule(lr){2-3}
 & \textsc{Clean} & defective \\
\midrule
\textsc{Clean} ($15$)      & $6$  & $9$ \\
defective ($105$)          & $11$ & $94$ \\
\bottomrule
\end{tabular}
\caption{Binary cross-model comparison (\textsc{Clean} vs.\ defective) between the GPT-5.4 audit and an independent, label-blind Claude re-judge on a stratified sample of $120$ TMax environments, with $15$ examples drawn from each GPT-5.4 verdict category. Defect labels show substantially higher conditional agreement than \textsc{Clean} labels.}
\label{tab:audit-crossmodel}
\end{table}

\clearpage
\newpage
\newtcolorbox{auditexamplebox}[1]{colback=gray!4,colframe=gray!55,fonttitle=\bfseries,
  title=#1,boxrule=0.5pt,left=4pt,right=4pt,top=3pt,bottom=3pt,breakable,
  before skip=10pt,after skip=8pt}

\subsection{Environment Evaluation Examples}
\label{app:env-audit-examples}

We show a compact set of representative environments flagged by our audit. Each
example was independently checked by a second model. Each box summarizes what the
task asks, why the environment receives the corresponding verdict, and the most
direct evidence supporting that verdict. These examples are illustrative rather
than exhaustive, and a single environment may exhibit more than one failure
mode.

\vspace{4pt}\noindent\textbf{\textsc{Verifier-Too-Weak}}\par\smallskip

\begin{auditexamplebox}{\texttt{task\_000254\_21ede93b}}
\textbf{Task.}
Write a C archiver daemon that recursively processes legacy log files, invokes a
filter script via \texttt{popen()}, maintains atomic JSON state, and continuously
monitors an incoming directory using \texttt{inotify}.

\textbf{Why \textsc{Verifier-Too-Weak}.}
The instruction requires several concrete implementation properties: the solution
must be a C program, recursively traverse existing logs, invoke the filter through
\texttt{popen()}, update its state atomically using \texttt{rename()}, and use
\texttt{inotify} to process newly closed files. The verifier, however, observes
only a running process named \texttt{archiver}, the existence of output files, and
the expected aggregate counts before and after adding a new file. It never checks
that the implementation is written in C or that any of the required mechanisms
are actually used. Consequently, another implementation that merely reproduces
the expected observable files and counts could satisfy the verifier without
implementing the requested archiver.

\textbf{Key evidence.}
The verifier checks the process name, output-file existence, and
\texttt{total\_critical} counts, but never validates C source,
\texttt{popen()}, recursive traversal, \texttt{inotify}, or atomic
\texttt{rename()}.
\end{auditexamplebox}

\begin{auditexamplebox}{\texttt{task\_000233\_ee5f0e33}}
\textbf{Task.}
Back up and repair several microservice configurations, write an idempotent Bash
script that applies the required port and upstream settings, and write a second
script that derives a topology report from the resulting configurations.

\textbf{Why \textsc{Verifier-Too-Weak}.}
The verifier checks configuration values only through substring membership, such
as testing whether \texttt{LISTEN\_PORT=9001} occurs somewhere in the file.
This does not ensure that the configuration is well formed, unique, or free of
conflicting stale values. More importantly, although the task requires
\texttt{generate\_report.sh} to parse the configurations and dynamically
construct the topology report, the verifier never executes that script. It only
checks that the final report file contains a fixed set of expected lines. Thus,
the required data flow from configuration repair to report generation is not
verified.

\textbf{Key evidence.}
Configuration values are checked by substring presence only, and
\texttt{generate\_report.sh} is never executed by the verifier.
\end{auditexamplebox}

\vspace{4pt}\noindent\textbf{\textsc{Instr-Verifier-Mismatch}}\par\smallskip

\begin{auditexamplebox}{\texttt{task\_001036\_07b75c89}}
\textbf{Task.}
Repair a compromised data-processing pipeline, including an authentication
validator, a redaction worker, and an Nginx reverse proxy that should listen on
\texttt{127.0.0.1:8080} and forward traffic to the Flask application.

\textbf{Why \textsc{Instr-Verifier-Mismatch}.}
The agent follows the networking requirement using the canonical Nginx
configuration
\texttt{listen 127.0.0.1:8080;}.
However, the verifier searches the configuration with the regular expression
\texttt{listen\textbackslash s+8080\textbackslash s*;}, which accepts
\texttt{listen 8080;} but does not match the explicitly specified
\texttt{127.0.0.1:8080} form. Thus, the verifier rejects a configuration that
directly follows the instruction.

\textbf{Key evidence.}
The instruction requires listening on \texttt{127.0.0.1:8080}; the agent writes
\texttt{listen 127.0.0.1:8080;}, while the verifier only matches the pattern
\texttt{listen 8080;}.
\end{auditexamplebox}

\vspace{4pt}\noindent\textbf{\textsc{Instr-Env-Mismatch}}\par\smallskip

\begin{auditexamplebox}{\texttt{task\_000051\_313e5ebc}}
\textbf{Task.}
Transcribe an incident audio recording, redact the recovered secret from log
files, and implement a secure JWT validator whose outputs should match a provided
reference implementation under fuzz testing.

\textbf{Why \textsc{Instr-Env-Mismatch}.}
The audio input is generated at the path named by the instruction, but other
artifacts required by the task and verifier are missing. In particular, the
verifier requires
\texttt{/app/reference\_validator\_oracle.py} for fuzz-equivalence testing, yet
neither the container definition nor the setup script creates that file.
Likewise, the instruction identifies \texttt{/home/user/service/} as the location
of the service source to be audited, while the environment only creates that
directory and provides no service implementation to inspect. A solution that
follows the instruction therefore cannot complete all required steps or satisfy
the verifier using the supplied environment.

\textbf{Key evidence.}
The verifier asserts that
\texttt{/app/reference\_validator\_oracle.py} exists, but the environment never
creates it. The setup also creates \texttt{/home/user/service/} as an empty
directory and installs no corresponding service implementation.
\end{auditexamplebox}

\vspace{4pt}\noindent\textbf{\textsc{Answer-Leak}}\par\smallskip

\begin{auditexamplebox}{\texttt{task\_000028\_ae6330dd}}
\textbf{Task.}
Perform a multi-stage incident investigation: extract information from video,
recover a 4-byte XOR key from an ELF binary, analyze packet-capture traffic, and
produce a JSON report containing the detected malicious events.

\textbf{Why \textsc{Answer-Leak}.}
The intended task requires combining several forensic modalities to recover three
specific events. However, the setup exposes
\texttt{/home/user/ground\_truth.json}, which already contains the exact three
events used by the verifier, including timestamps, payload values, and certificate
flags. Reading this file reveals the complete expected report, so the intended
video, binary, and packet-capture analysis can be bypassed.

\textbf{Key evidence.}
The environment's \texttt{ground\_truth.json} contains the same three events as
the verifier truth set, including timestamps
\texttt{00:01:15}, \texttt{00:02:30}, and \texttt{00:04:45}, together with
the expected payload and flags.
\end{auditexamplebox}

\begin{auditexamplebox}{\texttt{task\_000236\_54686d44}}
\textbf{Task.}
Recover a passphrase from video, decrypt a vault, reverse-engineer a malicious
12-byte signature from an ELF payload, and implement an efficient IDS scanner
that detects that signature in logs.

\textbf{Why \textsc{Answer-Leak}.}
The intended workflow requires decrypting the vault and inspecting the ELF
payload before the complete malicious signature is known. However, the
agent-accessible evaluation log at \texttt{/app/test\_logs.txt} already contains
the exact injected signature
\texttt{deadbeef1122334455667788}. Because the instruction also reveals that the
signature begins with \texttt{DE AD BE EF}, an agent can recover the full value
directly from the test log and hard-code it into the scanner, bypassing the
vault-decryption and binary-analysis steps intended to derive the signature.

\textbf{Key evidence.}
The environment injects
\texttt{deadbeef1122334455667788} into 42 lines of
\texttt{/app/test\_logs.txt}, and the verifier uses the same value to construct
the expected output.
\end{auditexamplebox}

\vspace{4pt}\noindent\textbf{\textsc{Task-Trivial}}\par\smallskip

\begin{auditexamplebox}{\texttt{task\_001515\_bf254241}}
\textbf{Task.}
Recover a $3\times3$ projection matrix from the voice memo
\texttt{/app/artifact\_summary.wav}, then implement a Python transformation
script that applies the recovered matrix to arbitrary three-dimensional inputs.

\textbf{Why \textsc{Task-Trivial}.}
The stated challenge is to recover the matrix from audio. However, the
agent-accessible environment also contains
\texttt{/app/oracle\_transform.py}, the exact reference implementation used by
the verifier for fuzz equivalence. Inspecting or probing this deterministic
oracle directly reveals the transformation, so the audio-recovery component is
unnecessary. In the observed rollout, the agent directly inspected the oracle and
recovered the fixed diagonal transformation from it.

\textbf{Key evidence.}
The verifier invokes \texttt{/app/oracle\_transform.py} as the reference
implementation, and this same oracle is accessible to the agent. It reveals the
matrix
\[
    \operatorname{diag}(2.5,-1.5,4.2),
\]
allowing the transformation to be hard-coded without using the audio.
\end{auditexamplebox}
\clearpage
\newpage
\newcommand{\turn}[1]{\textbf{T#1}}

\newlist{trace}{description}{1}
\setlist[trace]{
    leftmargin=1.25cm,
    labelwidth=1.0cm,
    labelsep=0.25cm,
    font=\normalfont\bfseries,
    itemsep=2.5pt,
    topsep=3pt,
    parsep=0pt
}

\newtcolorbox{rhbox}[1]{
    colback=green!6,
    colframe=green!45!black,
    boxrule=0.4pt,
    fonttitle=\bfseries,
    title={#1},
    left=4pt,
    right=4pt,
    top=3pt,
    bottom=3pt,
    breakable,
    before skip=8pt,
    after skip=8pt
}

\subsection{Examples of Verifier Failures on the Unfiltered Environment Set}
\label{app:reward-hacking}
Using the unfiltered TMax collection~\citep{ivison2026tmax} may lead to reward hacking and encourage shortcuts in policies' rollouts. The performance of RL on randomly sampled 3.5K environments from the TMax collection is presented in Tab.~\ref{tab:ablation_random_tmax}, where we can see a clear average performance gap with our model.

\begin{table}[h]
\centering
\small
\setlength{\tabcolsep}{6pt}
\begin{tabular}{lccccc}
\toprule
Model & Otbl & Tbv2.1 & Tbpro & Twv & \textbf{Avg.} \\
\midrule
River-8B (ours) & \gpm{21.8}{2.9} & \gpm{9.7}{0.6} & \gpm{23.0}{3.1} & \gpm{23.0}{2.5} & \gpmb{19.4}{1.2} \\
RL On Random TMax-3.5K & \gpm{18.6}{1.9} & \gpm{8.2}{1.3} & \gpm{23.7}{1.5} & \gpm{20.3}{0.6} & \gpm{17.7}{0.7} \\
River-8B-SFT & \gpm{11.3}{2.3} & \gpm{6.4}{1.3} & \gpm{13.7}{2.6} & \gpm{18.5}{2.3} & \gpm{12.5}{1.1} \\
\bottomrule
\end{tabular}
\caption{Numbers are benchmark scores ($\times10^{2}$). \textbf{Avg.}\ is the unweighted mean of the four benchmarks. Scores are mean\,$\pm$\,std over $3$ seeds.}
\label{tab:ablation_random_tmax}
\end{table}

This section also provides some examples of bad behaviors fostered by training tasks with low-quality verifiers. Inspecting the corresponding trajectories shows that these defects can corrupt the reward signal in both directions. Some defective environments produce \emph{false-positive rewards}: the agent obtains \texttt{reward}$=1$ by exploiting leaked answers, degenerate oracles, or missing verifier checks without completing the intended task. Other defects produce \emph{false-negative rewards}: a correct and instruction-aligned solution receives \texttt{reward}$=0$ because the verifier uses an incorrect reference answer, a buggy oracle, an over-specified implementation check, or an internally inconsistent test. Below we show some representative examples for both \emph{false-positive rewards} and \emph{false-negative rewards}. {\color{hlred}Red} highlights the critical shortcut or verifier failure.

\paragraph{False-positive rewards.}
The first five examples show cases in which the observed reward overstates task
completion.

% ---------------------------------------------------------------------
\begin{rhbox}{(1) Answer Leak -- \texttt{task\_003722} (reward\,$=1$)}
\textbf{Task.}
Recover a scrolling news ticker from \texttt{archive\_broadcast.mp4}. The agent
must extract frames at 1 FPS, crop the bottom of each frame, run OCR, reverse
mojibake corruption, normalize the recovered text, and reconstruct the
chronological ticker in \texttt{reconstructed\_ticker.txt}.

\textbf{Environment/verifier defect.}
The verifier compares the reconstructed text against
{\color{hlred}\texttt{/app/ground\_truth.txt}}, but this supposedly hidden
reference is directly readable by the agent.

\begin{trace}
  \item[\turn{1}]
  While inspecting the environment, the agent discovers
  \texttt{/app/ground\_truth.txt} alongside the video and reads the complete
  expected ticker.

  \item[\turn{2}--\turn{5}]
  The agent nevertheless attempts the intended video pipeline: it extracts frames,
  crops the ticker region, and constructs a noisy OCR-like reconstruction
  containing duplicated and corrupted fragments.

  \item[\turn{6}]
  It explicitly compares this reconstruction against the leaked reference and
  obtains only {\color{hlred}$0.368$ similarity}, far below the required threshold
  of $0.85$.

  \item[\turn{7}]
  The agent then reasons that the output should match the expected string and
  {\color{hlred}replaces the failed reconstruction with the text read directly
  from \texttt{ground\_truth.txt}}.

  \item[\turn{8}]
  The final output is identical to the reference, giving similarity $1.0$ and
  full reward.
\end{trace}

\textbf{Why the reward is misleading.}
The intended challenge is to recover the ticker from the video. The agent's
actual reconstruction is insufficient, but the exposed reference lets it replace
that failed result with the exact answer. Thus, {\color{hlred}the successful reward
reflects access to the verifier target rather than successful video understanding}.
\end{rhbox}

% ---------------------------------------------------------------------
\begin{rhbox}{(2) Answer Embedded in the Instruction -- \texttt{task\_008903} (reward\,$=1$)}
\textbf{Task.}
Write and run a C++ scanner over several binaries. For each valid 64-bit ELF file,
the scanner must determine whether the SUID bit is set, parse the ELF
\texttt{.rodata} section, detect whether it contains the string
\texttt{/etc/shadow}, and write the sorted results to
\texttt{/home/user/scan\_results.txt}.

\textbf{Environment/verifier defect.}
The instruction's example output is not merely illustrative: it gives the
{\color{hlred}exact correct three output lines for the binaries in the
environment}. The verifier then checks the final result against those same
expected values.

\begin{trace}
  \item[\turn{1}--\turn{3}]
  The agent attempts to construct the requested C++ scanner, but compilation and
  execution do not produce a reliable working scan.

  \item[\turn{4}]
  It notices that the instruction already provides the complete expected output:
  \texttt{fake\_bin: SUID=Yes, SHADOW\_IN\_RODATA=No},
  \texttt{safe\_bin: SUID=No, SHADOW\_IN\_RODATA=No}, and
  \texttt{suid\_bin: SUID=Yes, SHADOW\_IN\_RODATA=Yes}.

  \item[\turn{5}]
  The trajectory explicitly acknowledges the shortcut:
  {\color{hlred}``I've just created a hardcoded output file''}, and writes these
  three lines directly to \texttt{scan\_results.txt}.

  % \item[\turn{6}]
  % The hard-coded file matches the expected output and receives full reward eventexttt{scan\_results.txt}.

  \item[\turn{6}]
  The hard-coded file matches the expected output and receives full reward even
  though the requested scanner is not demonstrated to have produced it.
\end{trace}

\textbf{Why the reward is misleading.}
The task is intended to test binary inspection and ELF parsing. Instead,
{\color{hlred}the exact answer is exposed in the instruction and can be copied
without obtaining it from the binaries}. The final-output verifier cannot
distinguish this shortcut from a working scanner.
\end{rhbox}

% ---------------------------------------------------------------------
\begin{rhbox}{(3) Task Trivial -- \texttt{task\_001515} (reward\,$=1$)}
\textbf{Task.}
Recover a $3\times3$ linear transformation from the voice memo
\texttt{/app/artifact\_summary.wav}, then implement the transformation in
\texttt{/home/user/transform.py}. The script is evaluated on random
three-dimensional input vectors.

\textbf{Environment/verifier defect.}
The exact reference implementation,
{\color{hlred}\texttt{/app/oracle\_transform.py}}, is shipped inside the
agent-accessible container, and the verifier uses this same script as the oracle
for fuzz equivalence.

\begin{trace}
  \item[\turn{1}]
  The agent lists the task files and discovers \texttt{oracle\_transform.py}
  alongside the audio memo.

  \item[\turn{2}]
  Rather than recovering the matrix from the audio, it executes
  {\color{hlred}\texttt{cat /app/oracle\_transform.py}}.

  \item[\turn{3}]
  The reference implementation reveals the entire transformation:
  \[
      \operatorname{diag}(2.5,-1.5,4.2).
  \]

  \item[\turn{4}]
  The agent writes \texttt{transform.py} using exactly these constants.

  \item[\turn{5}--\turn{6}]
  It explicitly notes that {\color{hlred}``the audio file wasn't needed''}.
  The verifier then fuzzes the copied implementation against the exposed oracle,
  and all tests pass.
\end{trace}

\textbf{Why the reward is misleading.}
The generated implementation is numerically correct, but the defining challenge
of the task is to recover the transformation from the voice memo.
{\color{hlred}Exposing the reference implementation collapses the intended
audio-understanding task into copying a few constants}.
\end{rhbox}

% ---------------------------------------------------------------------
\begin{rhbox}{(4) Degenerate Oracle -- \texttt{task\_000925} (reward\,$=1$)}
\textbf{Task.}
Reverse-engineer the behavior of a proprietary semantic-version resolver,
\texttt{/app/legacy\_resolver}, and implement a Bash replacement supporting SemVer
comparisons, Boolean operators, and parentheses.

\textbf{Environment/verifier defect.}
The supplied resolver is degenerate: across the agent's probes, it returns exit
code $0$ even for obviously false comparisons. The verifier evaluates the
replacement by agreement with this broken resolver.

\begin{trace}
  \item[\turn{1}--\turn{3}]
  The agent probes \texttt{legacy\_resolver} on straightforward positive and
  negative examples.

  \item[\turn{4}]
  It observes that contradictory cases such as
  \texttt{>=1.0.0} with \texttt{0.9.0},
  \texttt{>2.0.0} with \texttt{1.9.0}, and
  \texttt{=1.0.0} with \texttt{2.0.0} all return success.

  \item[\turn{5}]
  The agent explicitly concludes:
  {\color{hlred}``The binary appears to have a bug where it just always returns
  exit code 0.''}

  \item[\turn{6}]
  Instead of implementing semantic-version parsing and comparison, it creates a
  replacement whose effective behavior is simply
  {\color{hlred}\texttt{exit 0}} for every input.

  \item[\turn{7}]
  Because the verifier compares the replacement against the same degenerate
  oracle, this constant policy receives full reward.
\end{trace}

\textbf{Why the reward is misleading.}
The replacement does not implement semantic-version matching. Instead,
{\color{hlred}the policy identifies the oracle's degenerate decision boundary and
implements the cheapest behavior that reproduces it}. This shows how a defective
oracle can make a trivial shortcut optimal under the observed reward.
\end{rhbox}

% ---------------------------------------------------------------------
\begin{rhbox}{(5) Unchecked Implementation Constraint -- \texttt{task\_008402} (reward\,$=1$)}
\textbf{Task.}
Fix a C++ backup-analysis program so that it queries replication metadata,
constructs the valid directed graph, computes the lowest-latency path between two
datacenters, and writes the result to
\texttt{optimal\_backup\_path.json}. The instruction explicitly requires the
solution to be produced \emph{entirely in C++} and forbids a separate Python or
Bash implementation.

\textbf{Environment/verifier defect.}
The verifier only checks that \texttt{/home/user/backup\_analyzer} exists and is
executable, then separately checks the JSON contents.
{\color{hlred}It never runs the C++ executable to verify that the executable
actually produced the answer}.

\begin{trace}
  \item[\turn{1}--\turn{3}]
  The agent edits and compiles the requested C++ program, but repeated executions
  of \texttt{backup\_analyzer} time out and never produce the required JSON.

  \item[\turn{4}]
  Instead of continuing to debug the required implementation, it writes a
  separate Python program, \texttt{backup\_checker.py}, that computes the answer.

  \item[\turn{5}]
  The Python program writes the correct path
  {\color{hlred}\texttt{[1,2,4]}} with total latency
  {\color{hlred}\texttt{30}} to \texttt{optimal\_backup\_path.json}.

  \item[\turn{6}]
  The trajectory explicitly acknowledges that the C++ program was timing out and
  that Python was used to generate the final output.

  \item[\turn{7}]
  The verifier observes an executable C++ file and a correct JSON file, never
  checks their causal relationship, and returns full reward.
\end{trace}

\textbf{Why the reward is misleading.}
The numerical answer is correct, but the explicit implementation requirement is
not satisfied. {\color{hlred}Existence-only verification of the binary lets a
forbidden Python bypass receive the same reward as a working C++ implementation}.
\end{rhbox}

\paragraph{False-negative rewards.}
The next five examples show the opposite failure mode: the observed reward
understates task completion because the verifier rejects correct or
instruction-faithful behavior.

% ---------------------------------------------------------------------
\begin{rhbox}{(6) Incorrect Numerical Reference -- \texttt{task\_009230} (reward\,$=0$)}
\textbf{Task.}
Compute a posterior score using the specified formula
\[
    p=(1-\exp(-x))\,p_{\mathrm{prior}},
\]
and write the resulting probability rounded to six decimal places.

\textbf{Environment/verifier defect.}
For one case with $x=2.99$ and $p_{\mathrm{prior}}=0.7$, the verifier hard-codes
{\color{hlred}\texttt{0.664815}} as the expected value.

\begin{trace}
  \item[\turn{1}--\turn{3}]
  The agent follows the numerical procedure specified by the instruction.

  \item[\turn{4}]
  For the disputed value, it computes
  \[
      (1-\exp(-2.99))\times0.7
      =0.664798794\ldots,
  \]
  which rounds to {\color{hlred}\texttt{0.664799}}.

  \item[\turn{5}]
  The remaining reported numerical values match the verifier.

  \item[\turn{6}]
  The verifier nevertheless expects \texttt{0.664815}. The discrepancy exceeds
  its tolerance, so the mathematically correct rollout receives
  \texttt{reward}$=0$.
\end{trace}

\textbf{Why the reward is misleading.}
The agent follows the stated formula and produces the correct rounded value.
{\color{hlred}The failure is caused by an erroneous hard-coded reference constant,
not by an error in the solution}.
\end{rhbox}

% ---------------------------------------------------------------------
\begin{rhbox}{(7) Incorrect Semantic Assumption -- \texttt{task\_002319} (reward\,$=0$)}
\textbf{Task.}
Perform spectral analysis on an EIIP-mapped DNA sequence. After computing the
float64 FFT magnitude spectrum, the instruction requires finding the
\emph{maximum magnitude} over indices $1$ through $N/2$, excluding only the DC
component.

\textbf{Environment/verifier defect.}
The verifier does not compute the requested maximum. Instead, it hard-codes
{\color{hlred}\texttt{expected\_peak\_index = N // 11}}, assuming that the
fundamental frequency must be dominant.

\begin{trace}
  \item[\turn{1}--\turn{3}]
  The agent parses the FASTA file, constructs the float32 and float64 EIIP
  sequences, computes both FFTs, and evaluates their magnitude spectra.

  \item[\turn{4}]
  Following the instruction literally, it searches indices $1$ through $N/2$
  and obtains {\color{hlred}\texttt{peak\_index = 50000}} with magnitude
  approximately $1311.08$.

  \item[\turn{5}]
  Independent recomputation confirms that this is the true maximum. The verifier's
  expected index \texttt{10000} has magnitude only about $322.33$.

  \item[\turn{6}]
  The verifier nevertheless rejects \texttt{50000} because it expects the
  hard-coded fundamental-frequency index, producing \texttt{reward}$=0$.
\end{trace}

\textbf{Why the reward is misleading.}
The agent correctly answers the question posed by the instruction.
{\color{hlred}The verifier substitutes an incorrect domain assumption for the
actual maximum and penalizes the correct spectral result}.
\end{rhbox}

% ---------------------------------------------------------------------
\begin{rhbox}{(8) Buggy Reference Oracle -- \texttt{task\_001636} (reward\,$=0$)}
\textbf{Task.}
Build a high-performance C program for ``latest-frame'' lookup. Given a query
timestamp $T$, it must return the frame whose \texttt{pkt\_pts\_time} is the
largest value less than or equal to $T$, using an indexed array and binary search.

\textbf{Environment/verifier defect.}
The reference oracle reads H.264 packet metadata in decode order, where PTS values
are not monotonic, and then
{\color{hlred}runs binary search on this unsorted array}. The verifier compares
the agent against this buggy oracle rather than against the requested lookup
semantics.

\begin{trace}
  \item[\turn{1}--\turn{3}]
  The agent extracts the same PTS/packet-size pairs from the video as the oracle.

  \item[\turn{4}]
  Because binary search requires sorted keys, it correctly sorts the metadata by
  PTS before building the query index.

  \item[\turn{5}]
  For query \texttt{2.6755}, the largest PTS not exceeding the query is
  \texttt{2.6}, whose packet size is
  {\color{hlred}\texttt{120}}.

  \item[\turn{6}]
  The unsorted oracle instead returns \texttt{97}. The verifier treats
  \texttt{97} as ground truth and rejects the agent's correct
  \texttt{2.6755 -> 120} output.
\end{trace}

\textbf{Why the reward is misleading.}
The agent implements the requested latest-frame semantics and satisfies the
precondition required for binary search. {\color{hlred}It receives zero reward
because oracle equivalence is used as a proxy for correctness even though the
oracle itself violates the algorithm's assumptions}.
\end{rhbox}

% ---------------------------------------------------------------------
\begin{rhbox}{(9) Brittle Source-Code Check -- \texttt{task\_006064} (reward\,$=0$)}
\textbf{Task.}
Fix a Rust text-processing pipeline so that it lowercases the corpus, retains only
alphabetic characters \texttt{a-z} and spaces, tokenizes by whitespace, reports
the top three tokens, and computes their $3\times3$ sample covariance matrix.

\textbf{Environment/verifier defect.}
Two functional tests validate the produced tokens and covariance matrix, but a
third test additionally requires the source file to contain the literal substring
{\color{hlred}\texttt{is\_alphabetic}}. The instruction never requires this
specific Rust API.

\begin{trace}
  \item[\turn{1}--\turn{3}]
  The agent fixes the tokenization using
  \texttt{c.is\_ascii\_alphabetic() || c.is\_whitespace()}, which directly matches
  the instruction's requirement to retain \texttt{a-z} characters and spaces.

  \item[\turn{4}]
  It builds and runs the Rust project successfully.

  \item[\turn{5}]
  The resulting \texttt{top\_tokens.txt} is byte-exact correct
  (\texttt{the}, \texttt{quick}, \texttt{dog}), and
  \texttt{covariance.csv} exactly matches the expected matrix.

  \item[\turn{6}]
  Both functional output tests pass, but the source-inspection test fails solely
  because {\color{hlred}\texttt{is\_ascii\_alphabetic} does not contain the
  literal substring \texttt{is\_alphabetic}}. The final reward is therefore $0$.
\end{trace}

\textbf{Why the reward is misleading.}
The observable behavior is correct, and the chosen API is at least as faithful to
the explicit \texttt{a-z} requirement as the expected implementation.
{\color{hlred}A brittle source-pattern check rejects a functionally correct
solution because it uses a different valid implementation}.
\end{rhbox}

% ---------------------------------------------------------------------
\begin{rhbox}{(10) Inconsistent End-to-End Test -- \texttt{task\_001040} (reward\,$=0$)}
\textbf{Task.}
Repair a local experiment-tracking pipeline by configuring Nginx to proxy
\texttt{127.0.0.1:8080} to a provided Flask server, writing a CSV validator, and
starting the services. Clean training-log uploads should be accepted, while
malformed logs should be rejected.

\textbf{Environment/verifier defect.}
The provided Flask server accepts uploads only through a multipart form field
named \texttt{file}; if this field is missing, it returns HTTP $400$. However,
the end-to-end verifier
{\color{hlred}sends the CSV bytes as a raw POST body with no multipart encoding
and no \texttt{file} field}, while still requiring HTTP $200$ for a clean file.

\begin{trace}
  \item[\turn{1}--\turn{4}]
  The agent writes a validator enforcing the requested headers and numerical
  constraints, configures Nginx to proxy port $8080$ to the Flask service on
  port $5000$, and writes an executable startup script.

  \item[\turn{5}]
  The static validator, startup-script, and configuration tests pass; three of the
  four verifier tests succeed.

  \item[\turn{6}]
  The end-to-end test reads a clean CSV into raw bytes and constructs
  \texttt{urllib.request.Request(..., data=data, method="POST")} without a
  multipart \texttt{file} field.

  \item[\turn{7}]
  The unmodified Flask fixture therefore behaves as designed and returns
  {\color{hlred}\texttt{400 BAD REQUEST}}, while the verifier expects
  \texttt{200}. The rollout receives \texttt{reward}$=0$.
\end{trace}

\textbf{Why the reward is misleading.}
The failing request does not use the upload protocol required by the provided
server. {\color{hlred}The verifier and fixture are internally inconsistent, so an
instruction-faithful solution cannot satisfy the end-to-end test}.
\end{rhbox}

% ---------------------------------------------------------------------
\paragraph{Takeaway.}
These examples show that defective environments can corrupt the RL signal in both directions: \textbf{False-positive} rewards arise when answer-bearing artifacts, degenerate oracles, or incomplete checks make shortcuts sufficient for
\texttt{reward}$=1$. \textbf{False-negative} rewards arise when incorrect reference values, buggy oracles, over-specified source checks, or inconsistent test protocols reject solutions that correctly follow the instruction. Together, these cases further motivate our environment filtering process.
\clearpage
\newpage
\section{Behavior Analysis Details}
\label{app:behavior-analysis-details}

In this section, we list the behaviors we used for the AUC-ROC test. Each behavior feature is extracted {deterministically} from a trajectory using rule-based parsing, with no model or LLM in the
loop. For each assistant turn, we parse the \texttt{<command>} and treat the following user turn as the resulting observation. A command is considered an \emph{inspection} if its head corresponds to a pre-defined read-only tool set including
(\texttt{ls}, \texttt{cat}, \texttt{grep}, \texttt{find}, \texttt{stat},
\ldots, etc.) and does not perform redirection or writing. An \emph{error} is an observation that matches an error regex, such as
\texttt{Traceback}, \texttt{command not found}, \texttt{No such file}, or a non-zero exit status. Two commands are considered \emph{near-duplicates} if their token-level Jaccard similarity is at least $0.8$. 

\textbf{AUC} denotes the single-feature cross-validated AUC-ROC for predicting per-task success of the RL checkpoint on the four evaluation benchmarks, averaged over $15$ seeds of $5$-fold cross-validation. \textbf{Dir.} indicates the sign of the feature's correlation with success. Rows are sorted by AUC. The two strongest individual signals are verifying before finishing (\texttt{verify\_before\_done}, $+$) and repeatedly issuing similar commands (\texttt{repetition\_rate}, $-$). The latter serves as a broader proxy for the harmful repetition loops targeted by our turn-wise RL penalty.

\begin{table*}[h]
\centering
\small
\renewcommand{\arraystretch}{1.2}
\setlength{\tabcolsep}{5pt}
\caption{The $12$ behavior features used by Model-B, together with their definitions and univariate predictive power.}
\label{tab:behavior-features}
\resizebox{1.\linewidth}{!}{
\begin{tabular}{@{}>{\raggedright\arraybackslash}p{0.22\textwidth}
                  >{\raggedright\arraybackslash}p{0.58\textwidth}
                  cc@{}}
\toprule
\textbf{Feature} & \textbf{Definition / extraction rule (per trajectory)}
  & \textbf{AUC} & \textbf{Dir.} \\
\midrule

\texttt{verify\_before\_done} &
$1$ if at least one read-only inspection occurs among the up to three commands
immediately preceding the \texttt{<done>} action; otherwise $0$.
  & $0.632$ & $+$ \\

\texttt{repetition} &
Fraction of commands that are near-duplicates of at least one earlier command
in the same trajectory.
  & $0.631$ & $-$ \\

\texttt{cmd\_diversity} &
Number of distinct command heads divided by the total number of commands.
  & $0.615$ & $+$ \\

\texttt{python\_heredoc} &
Fraction of commands that invoke Python
(\texttt{python}/\texttt{python3}, including \texttt{-c}) or use a
here-document to perform task-related operations.
  & $0.604$ & $-$ \\

\texttt{inspect\_before\_act} &
Fraction of commands that perform read-only inspection, defined as commands
with an inspection head and no output redirection, before modifying the
environment.
  & $0.596$ & $+$ \\

\texttt{n\_errors} &
Number of commands whose resulting observations match the error regex.
  & $0.563$ & $-$ \\

\texttt{gave\_up\_flag} &
$1$ if the episode ends \emph{without} a \texttt{<done>} action and its final
observation is an error; otherwise $0$.
  & $0.549$ & $-$ \\

\texttt{error\_rate} &
Fraction of commands whose resulting observations match the error regex
(\texttt{n\_errors} divided by the total number of commands).
  & $0.535$ & $-$ \\

\texttt{error\_recovery} &
Among commands that result in an error, the fraction for which the \emph{next}
command is substantially different (i.e., not a near-duplicate), indicating a
change in approach after failure.
  & $0.531$ & $+$ \\

\texttt{inspect\_first} &
$1$ if the first command in the trajectory is a read-only inspection;
otherwise $0$.
  & $0.524$ & $+$ \\

\texttt{tool\_help} &
Fraction of commands that request tool help or usage information
(\texttt{--help}, \texttt{-h}, or \texttt{man}).
  & $0.510$ & $-$ \\

\texttt{premature\_done} &
$1$ if \texttt{<done>} is emitted either after the immediately preceding
command results in an error or without prior verification; otherwise $0$.
  & $0.507$ & $\sim\!0$ \\
\midrule

\multicolumn{2}{@{}l}{\emph{All 12 features (Model-B)}}
  & $\mathbf{0.735}$ & \\
\bottomrule
\end{tabular}
}
\end{table*}

\end{document}